# Characterizing the vertical continuity of fuel strata based on TLS point clouds Gaussian decomposition and overlap metrics

**Jesús Torralba [1*], Luis Á. Ruiz [1], Pablo Crespo-Peremarch [1], Marina Simó-Martí [1], Juan Pedro Carbonell-Rivera [2]**

[1] Geo-Environmental Cartography and Remote Sensing Group (CGAT), Department of Cartographic Engineering, Geodesy and Photogrammetry, Universitat Politècnica de València, Valencia, Spain.

[2] Integrated Remote Sensing Studio, Department of Forest Resources Management, Faculty of Forestry, University of British Columbia, Vancouver Campus, Vancouver, BC, Canada.

* Correspondence: jetorpe @ upv.es

## Abstract

Vertical forest structure influences fire behavior by regulating understory-to-canopy fuel connectivity, yet operational fuel-type models represent this complexity through discrete classes that only indirectly capture vertical continuity. High-density point clouds offer an opportunity to characterize vertical fuel connectivity in a more explicit and continuous manner, but specific methodological approaches are still required. We applied Gaussian decomposition to vertical height profiles derived from terrestrial laser scanning (TLS) point clouds across 25 permanent forest plots to characterize geometric vertical fuel continuity. A multiscale analysis spanning from the plot scale to ~0.5 $m^2$ units was used to evaluate geometric, probabilistic, and overlap metrics for characterizing fuel strata. The method reproduced the vertical transition between shrub and tree strata, with median Pearson correlations ranging from 0.83 to 0.96 across analysis levels. Probability density overlap metrics and separability indicators, such as the Jeffries–Matusita distance, provided complementary descriptors of vertical continuity. Intermediate spatial scales (3–6 m cells) provided a useful compromise between spatial detail and the consistent representation of shrub-canopy interaction patterns. Finer scales increased variability due to point density reduction and profile irregularity, and low mean squared error values were found insufficient as sole performance indicators. Gaussian decomposition applied to TLS point clouds enables explicit quantification of vertical fuel continuity, a limitation of current discrete classification systems. Continuous vertical connectivity metrics derived from this approach offer a pathway for improving ladder fuel characterization and strengthening decision-making support in wildfire risk assessment.

## 1. Introduction

Forest structure is a fundamental attribute of forest ecosystems, describing the three-dimensional organization of vegetation elements, including their spatial position, size, density, and connectivity (McElhinny et al., 2005). This structural complexity influences a wide range of ecological and functional processes, from biogeochemical cycles and primary productivity (Gough et al., 2019; Satapathy and Dutta, 2025) to biodiversity patterns and species distribution (Bergen et al., 2007; Moudrý et al., 2023). In this regard, the relationship between plant diversity and ecosystem structural complexity appears to be almost universally positive (Coverdale and Davies, 2023). Beyond its ecological relevance, forest structure plays a central role in forest resource characterization (Fischer et al., 2019), forest management planning (Kacic et al., 2025), and the stability of managed forest systems (Liu et al., 2024). Numerous structural descriptors, such as canopy height and openness, tree density, the number of vegetation strata, and the vertical distribution of biomass, are commonly used to estimate key ecosystem attributes (Atkins et al., 2018).

In the context of wildfires, vegetation structure becomes even more critical, as the spatial distribution of fuels directly influences fire behavior (Agee, 1996; Huff et al., 1995). Fire spreads through available vegetative fuels, whose quantity, spatial arrangement, and connectivity determine both the rate of spread and fire intensity (Atchley et al., 2021; Parsons et al., 2017). In particular, dense fuel accumulation has been identified as a key driver of high-severity wildfires (Parks et al., 2018; Prichard et al., 2022). At the vertical level, vegetation organization regulates the interaction between understory and tree canopy fuels, influencing the likelihood of transitions from surface fires to crown fires (Forbes et al., 2022), where ladder fuels have been shown to predict wildfire severity more consistently than canopy fuel loads (Hakkenberg et al., 2024). Rodríguez y Silva et al. (2017) further showed that, without wind-driven convection, crown ignition required direct flame contact from an intense surface fire. In addition, litter and shrub fuel loads significantly influence surface fire behavior (Ottmar et al., 2012). Consequently, fuel continuity within the ecosystem determines the ability of fire to spread between vegetation components and persist across the landscape (Atchley et al., 2021).

Different Fire Behavior Fuel Types (FBFT), or fuel models, have been developed to simplify fuel structure into a limited set of representative classes (Rothermel, 1972; Scott and Burgan, 2005). They describe fuels using aggregated parameters, representing different vegetation types as discrete classes that integrate physical and structural fuel properties, and provide a robust operational framework for fire behavior simulation and wildfire risk assessment at the landscape scale (Aragoneses and Chuvieco, 2021; Domingo et al., 2020; Hoffrén et al., 2023). These models underpin widely used simulators such as FARSITE and FlamMap (Papadopoulos and Pavlidou, 2011) and serve as the basis for numerous decision-support systems for wildfire management. However, this simplification into discrete fuel classes results in a schematic representation of fuel structure, particularly regarding its vertical organization. While some models include parameters such as canopy base height or canopy fuel load, vertical continuity between strata is typically represented only indirectly. In structurally complex forests, where multiple strata coexist and overlap, this limitation constrains the accurate characterization of structural connectivity and fire-type transitions (Forbes et al., 2022; Torralba et al., 2024). Consequently, there is increasing interest in methodologies that provide a more explicit and detailed description of vertical fuel distribution and connectivity (Viedma et al., 2024). The growing availability of high-resolution three-dimensional data, particularly from Light Detection and Ranging (LiDAR) technologies, offers new opportunities to provide more continuous and quantitative representations of fuel structure.

Direct field-based quantification of vertical continuity between vegetation strata is challenging, requiring intensive sampling and often relying on indirect estimations (Kramer et al., 2014; Menning and Stephens, 2007; Ottmar et al., 2014). Common descriptors include canopy base height (CBH) and the fuel strata gap (FSG), defined as the distance between the understory and the canopy. While CBH captures only the tree layer, FSG explicitly accounts for both strata and enables the description of vertical fuel transitions (Cruz et al., 2004). Nevertheless, in field conditions, both metrics are typically derived from point-based measurements and provide a simplified representation of vertical structure, without explicitly capturing the continuous distribution of fuels. Advances in three-dimensional remote sensing have enabled more objective and scalable characterization of forest structure. Airborne LiDAR systems (ALS) have been widely used to estimate variables such as canopy height, CBH, canopy fuel load (CFL), and canopy bulk density (CBD) at stand and landscape scales (Andersen et al., 2005; Atkins et al., 2018; Erdody and Moskal, 2010; Fidalgo-González et al., 2019; Hermosilla et al., 2014; Kramer et al., 2014; Rocha et al., 2023). However, their ability to characterize understory remains constrained by canopy occlusion and limitations in point density, particularly in dense or structurally complex stands (Hancock et al., 2017). Moreover, these metrics primarily describe the position or separation between strata but provide limited information about the shape, extent, and degree of interaction between vertical fuel distributions. In the context of full-waveform LiDAR, additional approaches have been developed to retrieve structural information directly from the waveform signal. Among them, Gaussian decomposition is a key tool for separating overlapping echoes and retrieving structural information from the signal (Hancock et al., 2017; Zhou et al., 2017; Zhuang et al., 2015). Gaussian decomposition models a complex waveform as the sum of individual Gaussian components, each representing a distinct scattering response within the vegetation profile, thereby enabling the retrieval of structural information from overlapping signals. In this regard, Zhuang et al. (2015) developed metrics based on the position and area of Gaussian components derived from the Laser Vegetation Imaging Sensor (LVIS), demonstrating improved biomass estimation in temperate deciduous forests compared to conventional LiDAR metrics. Mountrakis and Li (2017) focused on optimizing signal fitting through hybrid linear-nonlinear iterative algorithms to enhance the detection of weak or overlapping Gaussian components, without explicitly addressing their structural interpretation. In contrast, Zhou et al. (2017) proposed an approach in which each Gaussian component is interpreted as a structural layer within the vertical profile, enabling the characterization of stratification through parameters such as mean height, dispersion, and amplitude. Nevertheless, in most of these studies, Gaussian decomposition has been primarily applied to ALS data in large stands, and oriented toward full-waveform processing, biomass estimation, or layer segmentation, rather than explicit, continuous, and interpretable measure of structural connectivity between the understory and the canopy. Terrestrial laser scanning (TLS) provides a much more detailed representation of vegetation, particularly in near-ground strata, due to its high point density and lateral acquisition geometry, which significantly reduces occlusion effects. This enables more accurate characterization of fuel vertical distribution and the estimation of metrics directly related to structural connectivity, including vertical density profiles, FSG, ladder fuel indicators (Flynn et al., 2023; García et al., 2011; Wilson et al., 2021; Yépez-Rincón et al., 2021), fine-scale 3D shrub and surface fuel mass characterization and structure (Rowell et al., 2020).

Vertical fuel continuity is especially relevant in Mediterranean ecosystems, where rural depopulation, socioeconomic changes, and decades of fire suppression have promoted progressive biomass accumulation and increased forest structural complexity (Chergui et al., 2018; Kreider et al., 2024; Pausas and Fernández-Muñoz, 2012). As a result, many Mediterranean landscapes exhibit high fuel loads distributed across multiple vertical strata, with dense shrub layers, tree regeneration, and mature canopies coexisting (Marino et al., 2018; Tanase et al., 2024; Torralba et al., 2018; Viedma et al., 2020). This configuration enhances the likelihood of high-intensity wildfires, particularly when vertical fuel connectivity links the understory and the canopy. Accordingly, characterizing the fuel structure, particularly shrublands, remains a key research priority in wildfire science (Little et al., 2025), with direct implications for fire behavior modeling and risk assessment (Ritter et al., 2023; Simó-Martí et al., 2025).

In this context, this study proposes a new approach to characterize the vertical continuity of forest structure from point clouds. The method is based on the application of Gaussian decomposition to height-distribution profiles derived from high-density TLS point clouds, from which metrics are extracted to quantify the degree of overlap and vertical connectivity between tree canopy and understory vegetation strata. This approach is evaluated in a Mediterranean forest ecosystem characterized by high structural heterogeneity and the coexistence of two main fuel strata: shrub vegetation and the tree canopy. The study pursues three main objectives: (1) to develop a method for the characterization of the vertical forest structure based on the Gaussian decomposition of TLS-derived vertical profiles; (2) to propose and evaluate metrics allowing us to quantify the overlapping degree between shrub and tree strata; and (3) to evaluate how the scale factor influences the characterization of vertical fuel continuity. This work contributes to the development of quantitative methodologies for describing forest vertical structure and connectivity, providing new tools for fuel characterization and improving the understanding of structural conditions potentially contributing to vertical fire propagation.

## 2. Materials and Methods

### *2.1. Study area*

The study area (Figure 1a) is located in the Mediterranean forest of the Sierra de Espadán Natural Park (Castellón, Spain), in the eastern Iberian Peninsula. This area is representative of the climatic conditions of the western Mediterranean, characterized by mild and low-precipitation winters and warm, dry summers.

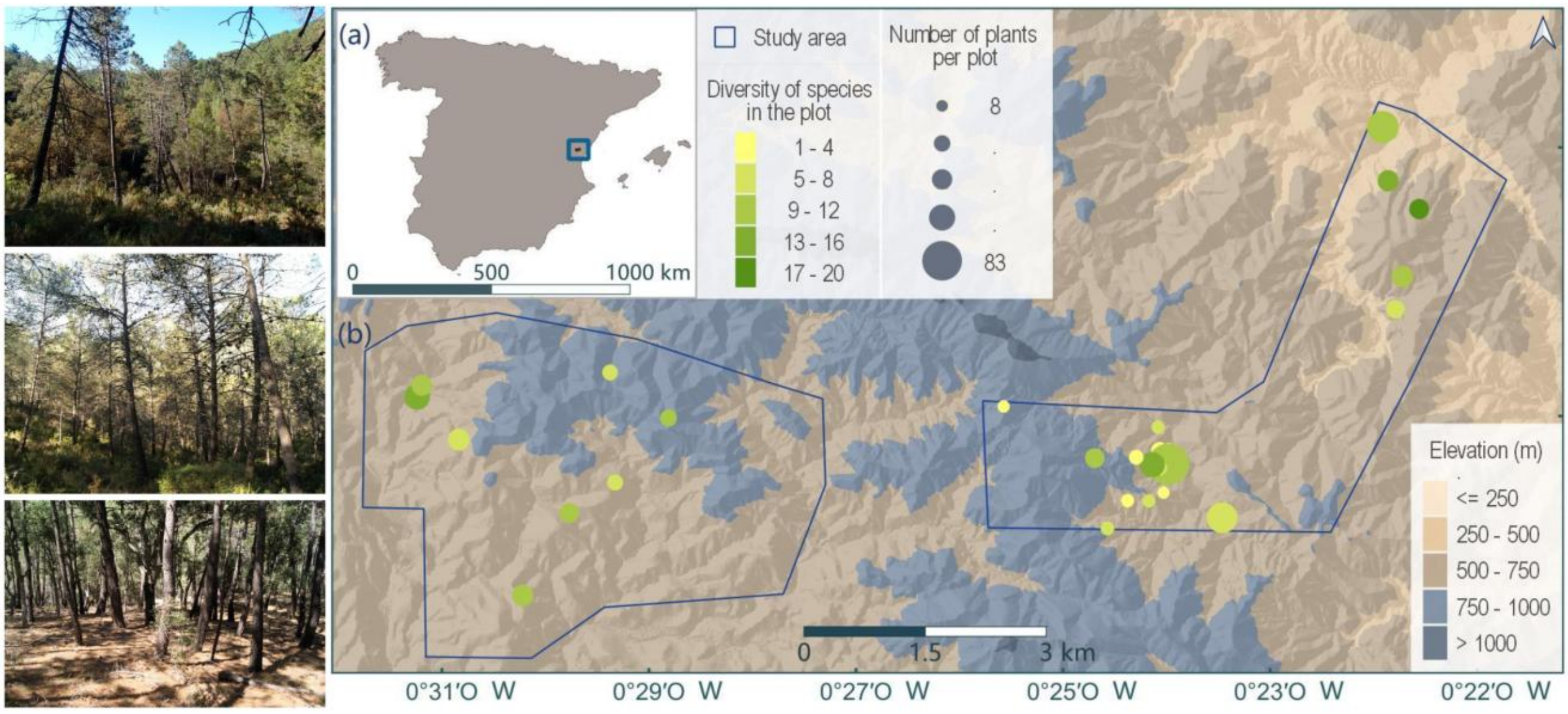

**Figure 1.** Study area and ground photographs of plots dominated by *Pinus pinaster*, *P. halepensis* and a mixed plot of *P. pinaster* and *Quercus suber* (from top to bottom). (a) Location of the study area in eastern Spain. (b) Spatial distribution of the plots in the Sierra de Espadán, where plot size is proportional to the number of sampled plants and the color scale (from yellow to dark green) represents species diversity.

The vegetation is composed of heterogeneous, predominantly even-aged forest stands, in which the tree layer is dominated by young to mature stands of *Pinus halepensis* Mill. and *P. pinaster* Aiton, with scattered codominant individuals of *Quercus suber* L. and *Q. ilex* L. (Figure 2a). This structural pattern reflects a strong dominance of pine-dominated stands, likely linked to past reforestation practices, resulting in a limited occurrence of mixed pine-oak stands. Even within these mixed formations, *P. pinaster* and *P. halepensis* dominate both stem density and basal area, whereas oak species play a secondary structural role (Torralba et al., 2018). Spatial variability in tree density and canopy cover fraction has led to pronounced differences in the distribution, density, and diversity of shrub species. The shrub layer is dominated by species such as *Genista scorpius* (L.) DC., *Salvia rosmarinus* Spenn., *Rhamnus alaternus* L., *Cistus albidus* L., *Pistacia lentiscus* L., *Quercus coccifera* L., *Erica arborea* L., and *Juniperus oxycedrus* L. (Figure 2b; Figure A1).

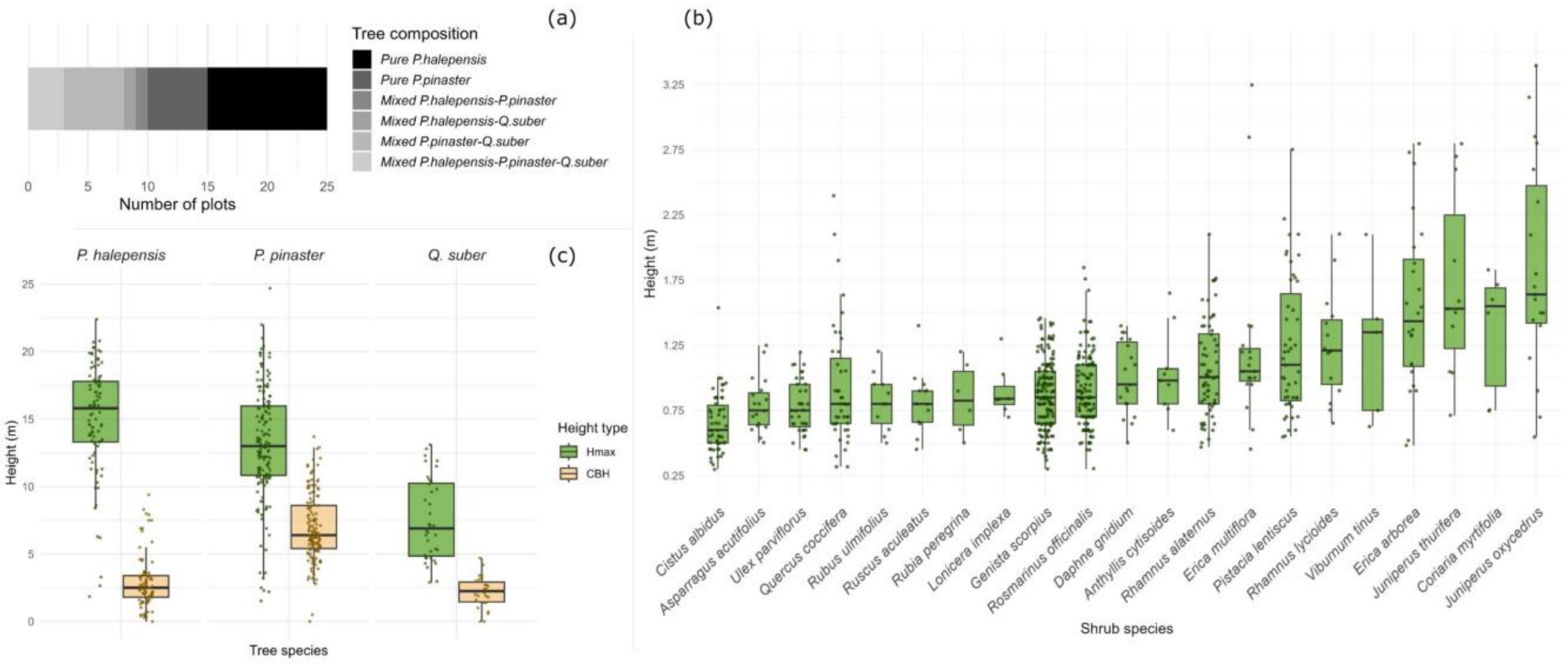

**Figure 2.** Tree species composition and height distribution of tree and shrub species. (a) Dominant tree species in the sampling plots. (b) Distribution of maximum height (Hmax) of the main shrub species present in the plots (species represented by fewer than six individuals were excluded, and one *Erica arborea* individual with a maximum height of 3.52 m was not displayed). Species are ordered by median Hmax. Regenerated tree species (DBH < 5 cm) were absent from the sampled plots. (c) Distribution of maximum height (Hmax) and canopy base height (CBH) of the tree species present in the plots.

The study comprised 25 permanent circular plots, each with a radius of 6 m (113 m²), surveyed in 2022. Plot elevation ranged from 375 to 877 m a.s.l. (Figure 1b), while the slope varied approximately between 4.8° and 24.9°. The study plots were characterized by two main fuel strata: a shrub layer and an overstory tree layer. Tree regeneration (DBH < 5 cm) was absent from the sampled plots and therefore did not constitute an additional intermediate stratum. The vertical organization of fuels showed a clear dependence on the composition of the dominant tree layer, with lower mean CBH in plots dominated by *P. halepensis* (2.93 ± 1.95 m) and higher CBH values in plots dominated by *P. pinaster* (6.92 ± 2.51 m). In mixed plots with the presence of *Q. suber*, the CBH values were lower (2.18 ± 1.17 m) (Figure 2c). In the shrub layer, the mean maximum height ranged from 0.52 to 4.15 m (Figure 2b), with an average crown cover of less than 1 m². However, some species, such as *Erica arborea*, *Erica multiflora* L., and *Juniperus oxycedrus*, exceeded crown cover values of 1.40 m², reflecting a high structural heterogeneity.

## 2.2. TLS data acquisition

The TLS data were acquired using a phase-based Faro Focus 3D 120 laser scanner (Faro Technologies Inc., Lake Mary, FL, USA). Five scan positions were established in each plot: one central position and four additional positions located 7.5 m away in the cardinal directions, with the aim of minimizing occlusions and adequately capturing the structural variability of the fuel complex (Torralba et al., 2022). Scans were acquired using a 1/5 resolution and a 4× quality setting, resulting in an approximate acquisition time of 6 minutes per scan position. Scan registration was performed using reference spheres whose positions were determined through conventional topographic surveying, ensuring the visibility of at least four spheres from each scan position.

### *2.3. Methodology*

A general overview of the methodological workflow is shown in Figure 3. First, TLS data were acquired in the field using a multi-scan configuration per plot. The resulting point clouds were then processed through noise filtering and scan registration. Once registered, the point clouds were normalized with respect to the ground surface by generating a plot-specific digital terrain model (DTM). To avoid artifacts in the characterization of vertical fuel continuity, points corresponding to tree stems were identified and removed. Each plot was subsequently subdivided hierarchically into spatial units of different sizes, defining five analysis levels that enabled evaluation of vertical structure across multiple spatial scales. For each spatial unit, histograms of normalized heights were constructed, and an optimal bin size was determined, after which a common bin width was imposed per level to ensure intralevel comparability. The discrete histograms were then converted into continuous profiles and smoothed using a Savitzky–Golay filter, followed by a Gaussian decomposition procedure based on peak detection and nonlinear fitting of Gaussian curves, allowing for the identification of up to four structural components per vertical profile. To reinforce structural coherence across scales, the search for Gaussian curves at finer levels was constrained using intervals inherited from the immediately coarser levels. From the fitted curves, distance, separability, and overlap metrics were computed between adjacent curve pairs to characterize the vertical organization of the profiles. Finally, a reference segmentation independent of the Gaussian decomposition procedure was generated by segmenting the preprocessed TLS point clouds into shrub and tree strata and applying the same multiscale spatial scheme and vertical discretization used in the Gaussian analysis. The observed proportions of each stratum along the vertical profile were then compared with the proportions derived from the Gaussian curves within the overlap intervals between strata. Agreement between both representations was evaluated using Pearson correlations and bin-wise error analysis. File-level agreement was further assessed using the weighted mean squared error (WMSE) and the weighted root mean squared error (WRMSE).

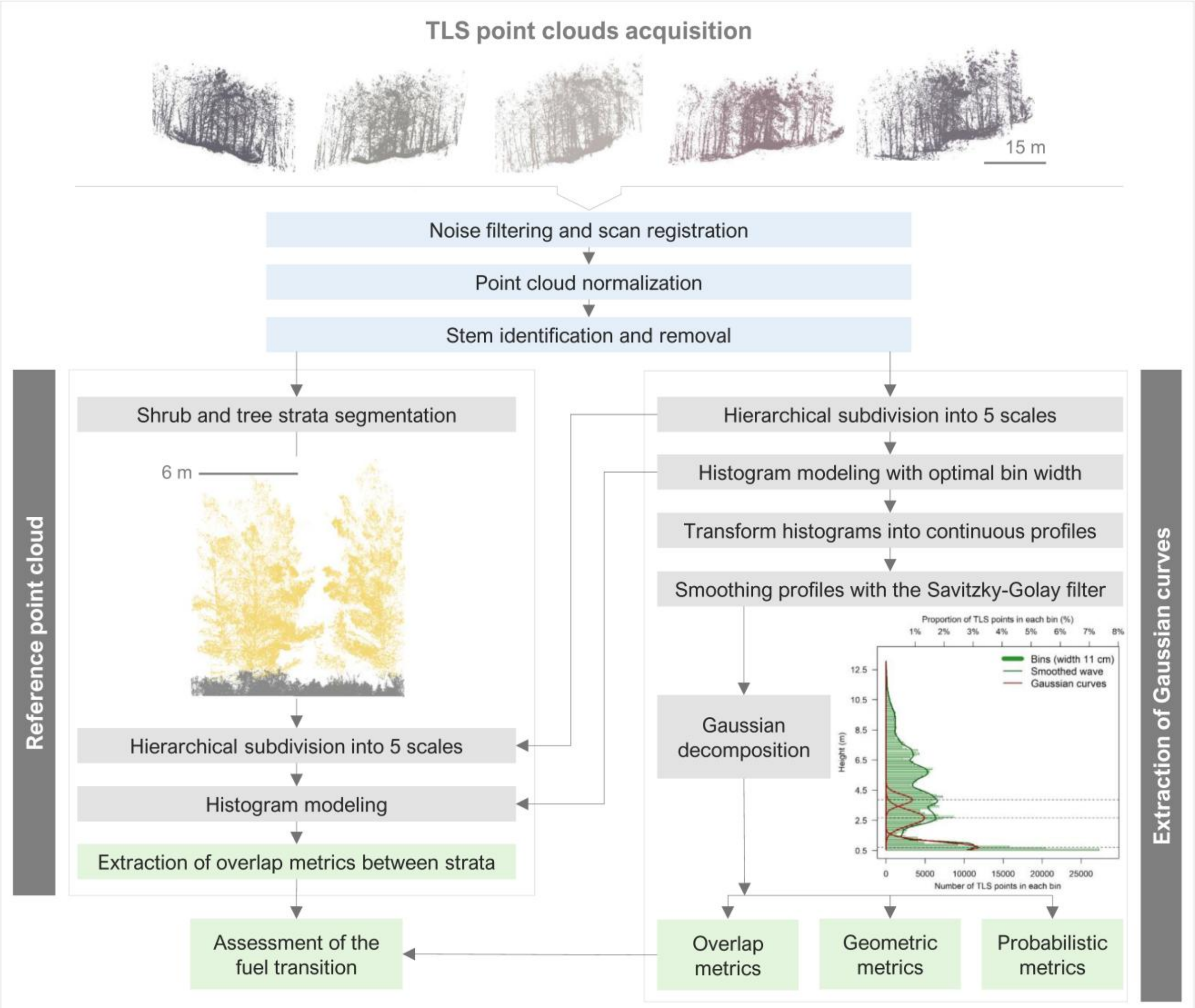

**Figure 3.** Schematic representation of the methodological workflow. Blue boxes represent point cloud pre-processing, grey boxes represent the processing steps, and green boxes represent the evaluation metrics. In the diagram corresponding to the "Reference point cloud" section, the beige-yellow color represents the tree stratum, and the brownish-grey color represents the shrub stratum. In the diagram corresponding to the "Extraction of Gaussian curves" section, the light-green bars represent the height histogram (binning), the dark-green line represents the smoothed profile, and the resulting Gaussian curves are shown in red. Dashed lines indicate the mean height ($\mu$, in meters) of each Gaussian component.

### 2.3.1. TLS data processing

Prior to registration, a supervised removal of outliers and noise was performed on the TLS point clouds. Co-registration between scans was carried out using a semi-supervised, object-based approach, selecting at least four reference spheres visible in adjacent scans. The mean registration error per plot was 6 ± 2 mm. Both noise removal and the registration procedures were performed using Trimble RealWorks© v11.3 (Sunnyvale, CA, USA).

Point clouds were normalized with respect to the ground surface. To this end, TLS data were classified into ground and vegetation points using LAStools® v230212. Points classified as ground were interpolated to generate a continuous surface corresponding to the DTM, which was subsequently used to normalize point-cloud heights. The average number of points per plot, excluding ground points, was 18,318 ± 10,261, with a maximum of 48,142 and a minimum of 9,314. Tree stems were removed prior to the analysis, as they correspond to heavy woody fuels (1000-h time-lag class), which play a limited role in vertical spread and are not directly involved in the propagation of flames between surface and canopy fuels. In addition, their inclusion could artificially lead to an overestimation of vertical fuel continuity. This filtering was performed using the expert model of the 3Dfin plugin (Laino et al., 2024) implemented in (CloudCompare, 2024).

2.3.2. Extraction of Gaussian curves

The vertical structure was analyzed hierarchically across five spatial scales. Level 1 corresponded to the entire plot (113 $m^2$), whereas subsequent levels were obtained through progressive spatial subdivisions, reaching analysis units of approximately 0.56 $m^2$ at Level 5 (Figure 4a and 4b). At Levels 4 and 5, peripheral columns with extremely small surface areas were discarded to ensure structural representativeness of the vertical profiles. This multiscale framework enabled assessment of the stability of structural descriptors as a function of spatial resolution, increasing the theoretical number of derived spatial units from 25 units at Level 1 to approximately 6,400 units at Level 5. These progressively subdivided units were used to assess sensitivity to spatial scale.

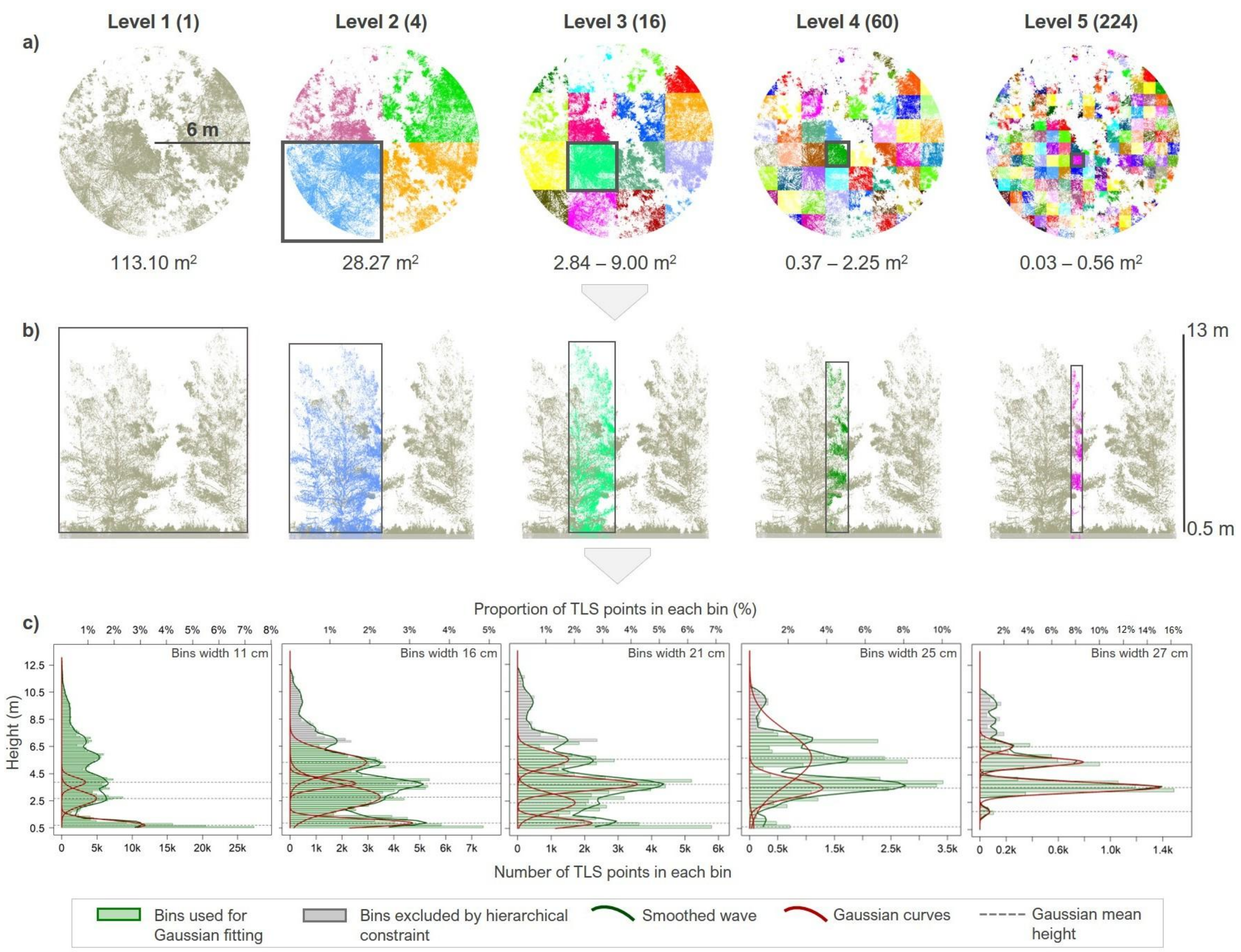

**Figure 4.** (a) Spatial scales used for Gaussian curve extraction. Level 1 corresponds to the entire plot (113.10 m²). Level 2 represents four quadrants (28.27 m² each). Level 3 subdivides each quadrant into four units ranging from 2.84 to 9.00 m². Level 4 subdivides each subquadrant into four units ranging from 0.37 to 2.25 m² (64 theoretical subdivisions per plot, of which 60 were retained for analysis). Level 5 further subdivides into units ranging from 0.03 to 0.56 m² (256 theoretical subdivisions per plot, of which 224 were retained for analysis). The number of subdivisions per plot is indicated in parentheses. Colors are arbitrary. (b) Vertical representation of the TLS point cloud. The full plot is shown in light grey, while the colored areas represent the analysis column at each spatial level (Levels 2–5). (c) Gaussian decomposition of vertical profiles. Light green bars represent the height histogram (bins) used for Gaussian fitting, while grey bars indicate bins excluded by the hierarchical constraint. The smoothed wave is shown in dark green, and the fitted Gaussian curves are shown in red. Dashed horizontal lines represent the mean height of each Gaussian component.

For each spatial unit, a histogram of normalized heights was constructed, excluding returns below 0.5 m to minimize the influence of ground and herbaceous vegetation (Figure 4c). The optimal vertical bin size was determined using Knuth's optBINS algorithm (Knuth, 2019), which estimates the number of bins under a fully Bayesian framework by modeling the histogram as a piecewise-constant density estimator. Unlike empirical rules (e.g., Scott or Freedman–Diaconis), this method selects the partition that maximizes the posterior probability, providing an objective discretization that adapts to the intrinsic structure of the data. To ensure intralevel comparability, a common bin width was imposed per level. First, the optimal bin width was estimated for each spatial units (after applying the 0.5 m height threshold). The median bin width per level was then computed and applied uniformly to all files within that level. The final bin widths were 0.11 m (Level 1), 0.16 m (Level 2), 0.21 m (Level 3), 0.25 m (Level 4), and 0.27 m (Level 5). Spatial units with insufficient point density (< 10 points) were discarded; specifically, 0.3% of the spatial units at Level 3, 1.1% at Level 4, and 3.5% at Level 5 were removed.

Discrete histograms (bin frequencies) were transformed into continuous profiles and smoothed using a Savitzky–Golay filter (Savitzky and Golay, 1964) (Figure 4c). This filter locally fits a polynomial over a moving window, reducing noise while preserving the overall curve shape and the location of local maxima. Combinations of polynomial order ($p$) and window size ($n$) were evaluated to assess robustness against over-smoothing or spurious peak detection (Figure A2). Specifically, $p$ values in {2, 3} and window sizes $n$ in {5, 7, 9, 11, 13} were tested, ensuring that $n$ remained odd. Based on this evaluation, a fixed polynomial order ($p = 2$) was selected, as it provided stable smoothing while preserving the overall signal shape, particularly in profiles with low point density. Window size was adjusted according to the spatial scale, using larger windows at coarser levels and smaller windows at finer resolutions to maintain the representativeness of the Gaussian components. The final configuration consisted of $n = 11$ at Level 1, $n = 9$ at Levels 2–4, and $n = 7$ at Level 5, balancing the detection of multimodal structures and the preservation of fine-scale vertical detail.

Following smoothing, a Gaussian decomposition procedure inspired by Zhou et al. (2017) was applied, implemented in R through a workflow combining peak detection and nonlinear fitting. Local maxima were identified using a centered three-bin neighborhood, and a maximum was retained as a candidate peak only when its magnitude exceeded a relative threshold defined with respect to the maximum value of the smoothed profile. The threshold was evaluated sequentially from 1.00 to 0.10 in decrements of 0.01, resulting in 91 configurations and allowing the number of detected peaks to adapt to the structural complexity of each profile.

For each detected peak configuration, a mixture of independent Gaussian components was fitted. Each Gaussian component $g_i(x)$ was:

$$g_i(x) = A_i \cdot \exp\left(-\frac{(x-\mu_i)^2}{2\sigma_i^2}\right) \tag{1}$$

parameterized by amplitude ($A_i$), mean ($\mu_i$), and standard deviation ($\sigma_i$), where $\mu$ represents the characteristic peak height and $\sigma_i$ its vertical dispersion. Initial values for $A$, $\mu$, and $\sigma$ were derived directly from the detected peaks using the expressions provided in Appendix B (Eqs. B1-B4). These starting values were subsequently optimized by nonlinear least squares using the Levenberg–Marquardt algorithm implemented in the R package *minpack.lm* (Elzhov et al., 2023), selected for its robustness under to approximate initial conditions and its stability in the presence of noisy signals. For each Gaussian, standard errors of $A$, $\mu$, and $\sigma$ were derived from the covariance matrix of the fitted parameters, computed from the Jacobian evaluated at convergence and the residual variance. To characterize the vertical extent of each Gaussian component, complementary descriptors were calculated: i) classical inflection-based intervals defined as $\mu \pm \sigma$ (and multiples thereof), and ii) coverage intervals associated with predefined fractions of the area under the curve. Specifically, parametric bands of the form $\mu \pm z\sigma$ were computed for coverage levels ranging from 50% to 97.5% (increments of 2.5%), where:

$$z = \Phi^{-1}\left(\frac{1 + coverage}{2}\right) \tag{2}$$

and $\Phi^{-1}$ denotes the inverse standard normal cumulative distribution function. In parallel, empirical intervals were derived directly from the smoothed profile, including: a) symmetric intervals around $\mu$ accumulating a specified fraction of the total area, and b) the shortest contiguous interval containing the target coverage fraction, equivalent to a Highest Density Interval (HDI).

The procedure was executed iteratively for each spatial unit and parameter combination ($p$, $n$, threshold). A configuration was considered valid when the nonlinear fit was completed without error and returned no missing parameter estimates, with positive $A_i$, $\mu_i$, and $\sigma_i$ for every component. Configurations producing fitting errors or invalid parameters were rejected, and the search continued with the next threshold. Candidate solutions were evaluated hierarchically, prioritizing a valid four-component solution, followed by three, and, as a last resort, two components (Figure 4.c). The threshold configurations were examined in a predefined order from highest to lowest; the search stopped when a valid four-component solution was found, while the last valid three- and two-component solutions encountered were retained as fallbacks. Selection was therefore based on fit validity, component number, and the predefined search sequence.

When the initial threshold range did not provide a valid four-component solution, a more permissive search from 0.90 to 0.02 was performed. If no valid solution containing at least two components was obtained after completing these steps, the profile was excluded from the Gaussian-based analyses. The retained components were limited to a maximum of four and ordered by increasing peak height ($\mu$), defining a strictly geometric indexing scheme (Gaussian 1–4).

To reinforce multiscale consistency and reduce spurious solutions at finer resolutions, a sequential constraint was applied. The allowable vertical search range for Gaussian fitting at a given level was restricted using information from the immediately coarser level (e.g., Levels 2→3, 3→4, and 4→5). Before defining the search intervals for the child units, each parent-level Gaussian component was screened for excessive vertical dispersion. Its effective width was defined as $8\sigma$, corresponding to the interval from $\mu - 4\sigma$ to $\mu + 4\sigma$, and was compared with the total vertical span of the parent profile, calculated as the difference between its maximum and minimum bin heights. If $8\sigma$ equaled or exceeded this span, the component was excluded from inheritance because it would not provide a useful restriction on the child-level search domain. This exclusion affected only the inheritance step and did not remove the component from the parent-level decomposition. For the remaining components, the inherited search intervals were defined as follows:

$$[L, U] = [\mu - k\sigma, \mu + k\sigma] \quad (3)$$

where $L$ and $U$ denote the lower and upper vertical bounds of the inherited search interval, and $k$ = 2 by default. The choice of $k$=2 corresponds to approximately 95.44% of the Gaussian area under a normal distribution, thereby capturing the dominant vertical mass of each component while limiting excessive vertical expansion.

To avoid imposing an excessively narrow search range on the child units, and based on the preliminary tests of the sequential procedure, inherited intervals narrower than 0.50 m were considered too restrictive and were discarded before inheritance. The remaining intervals were expanded by 0.25 m at both limits to accommodate local geometric differences among subdivisions. Overlapping intervals were first merged; when separated intervals ("islands") remained, their minimum-to-maximum envelope defined the final search domain. Finally, the child-level profile was masked to retain only bins within the inherited intervals, and peak detection and Gaussian fitting were repeated within this constrained vertical domain. When no inherited interval was available, the complete child-level profile was retained. This hierarchical restriction reduces the emergence of spurious peaks in small spatial units and enforces structural coherence across scales. Processing availability was tracked separately for each analysis level. Three successive sets of spatial units were distinguished: the theoretical units generated by spatial subdivision, the units containing vegetation points after subdivision, and the units retained after vertical binning (Table A1).

2.3.3. Gaussian pairwise distance and overlap metrics

Once Gaussian components were extracted and ordered by increasing peak height ($\mu$), pairwise comparisons were performed between adjacent components (Gaussian 1–2, 2–3, and 3–4). At this stage, no ecological interpretation was imposed; comparisons were conducted purely on the basis of the geometric and statistical properties of the fitted Gaussian curves. To characterize complementary aspects of vertical organization, three families of metrics were computed. Geometric metrics quantify the spatial configuration of the components, including peak separation, intersection heights, and interval overlap. Probabilistic metrics evaluate the statistical separability between Gaussian distributions using distance and information-based measures. Finally, overlap metrics quantify the degree of vertical interaction between components by measuring the shared area between their curves or the shared probability mass between their corresponding probability density functions. The complete set of metrics is summarized in Table 1.

**Table 1.** Summary of pairwise Gaussian distance and overlap metrics used to quantify vertical separation, statistical discriminability, and structural continuity between Gaussian components. Equations are given for two components $g_1$ and $g_2$, parameterized by amplitude ($A$), mean ($\mu$), and standard deviation ($\sigma$).

| Metric Type | Metric Name | Abbr. | Measure | Equation | Range | Reference |
|---|---|---|---|---|---|---|
| Geometric | Absolute peak separation | \|ΔH\| | Vertical distance between peak positions (m). | $\lvert\mu_1 - \mu_2\rvert$ | [0,∞) | |
| | Lower inter-section height | Xcross1 | Lower height at which two Gaussian curves intersect (structural boundary; m). | $Xcross_{1,2} = \frac{-b \pm \sqrt{b^2 - 4ac}}{2a}$ ; $a = \frac{1}{\sigma_2^2} - \frac{1}{\sigma_1^2}$ ; $b = -\frac{2\mu_2}{\sigma_2^2} + \frac{2\mu_1}{\sigma_1^2}$ ; $c = \frac{\mu_2^2}{\sigma_2^2} - \frac{\mu_1^2}{\sigma_1^2} - 2\ln\left(\frac{A_2}{A_1}\right)$ | (−∞,∞) | |
| | Upper inter-section height | Xcross2 | Upper height at which two Gaussian curves intersect (structural boundary; m). | | | |
| | Intersection-over-Union of parametric in-tervals | IoU(z) | Geometric overlap between parametric vertical intervals (independent of amplitude). | $IoU(z) = \frac{\lvert I_1 \cap I_2\rvert}{\lvert I_1 \cup I_2\rvert}$ ; $I_i(z) = [\mu_i - z\sigma_i, \mu_i + z\sigma_i]$ | [0,1] | |
| Probabilistic | d-prime | d′ | Standardized separation be-tween Gaussian peaks rela-tive to their pooled disper-sion. | $\frac{\lvert\mu_1 - \mu_2\rvert}{\sqrt{\frac{\sigma_1^2 + \sigma_2^2}{2}}}$ | [0,∞) | (Macmillan and Creel-man, 1991) |
| | Jeffries–Matusita | JM | Bounded separability index derived from the Bhattacharyya distance (Bd). | $JM = \sqrt{2(1 - e^{-Bd})}$ ; $Bd = \frac{1}{2}\ln\left(\frac{\sigma_1^2 + \sigma_2^2}{2\sigma_1\sigma_2}\right) + \frac{(\mu_1 - \mu_2)^2}{4(\sigma_1^2 + \sigma_2^2)}$ | [0,√2] | (Bhattachar yya, 1943; Swain and King, 1973) |
| Overlap | Distribution overlap coef-ficient | $OVL_{PDF}$ | Overlap between the proba-bility density functions of two Gaussian distributions, measuring their shared probability mass inde-pendently of amplitude. | $\int \min\ (f_1(x), f_2(x))\ dx$ | [0,1] | (Inman and Bradley, 1989) |
| | Directional overlap frac-tions | $OVL_{tree}$ | Fraction of the tree Gauss-ian component that overlaps with the shrub component. | $\frac{\int \min\ (g_2, g_1)\, dx}{\int g_2\, dx}$ | [0,1] | |

*In $OVL_{PDF}$, $f_i(x)$ denote normalized Gaussian probability density functions.

### 2.3.4. Validation

Validation was based on comparing the observed vertical overlap between strata with the overlap modeled through Gaussian decomposition. This analysis was applied independently at each of the five spatial levels defined in the multiscale framework to examine the stability of the structural representation as a function of spatial scale. In this way, we assessed whether the geometric separation captured by the Gaussian components coherently reproduces the real structural overlap between the shrub and tree strata, a key aspect for characterizing vertical fuel continuity.

To obtain a reference segmentation independent of the Gaussian fitting procedure, the TLS point clouds previously processed as described in Section 2.3.1 (i.e., normalized to the ground surface, noise-filtered, and with tree stems removed) were segmented. This procedure was applied to the 25 plots. Each point cloud was segmented into two strata (Figure 5a): shrub and tree. Segmentation was performed in CloudCompare with expert manual editing using a combination of scalar field-based classification tools (e.g., normalized height and intensity), visual inspection of the images acquired with the TLS scanner, and field photographs. These complementary sources facilitated shrub-tree boundary identification and ensured a coherent and consistent assignment of points throughout the entire vertical profile. Gaussian curves, fitted parameters, and derived overlap metrics were not used to assign points to the reference strata.

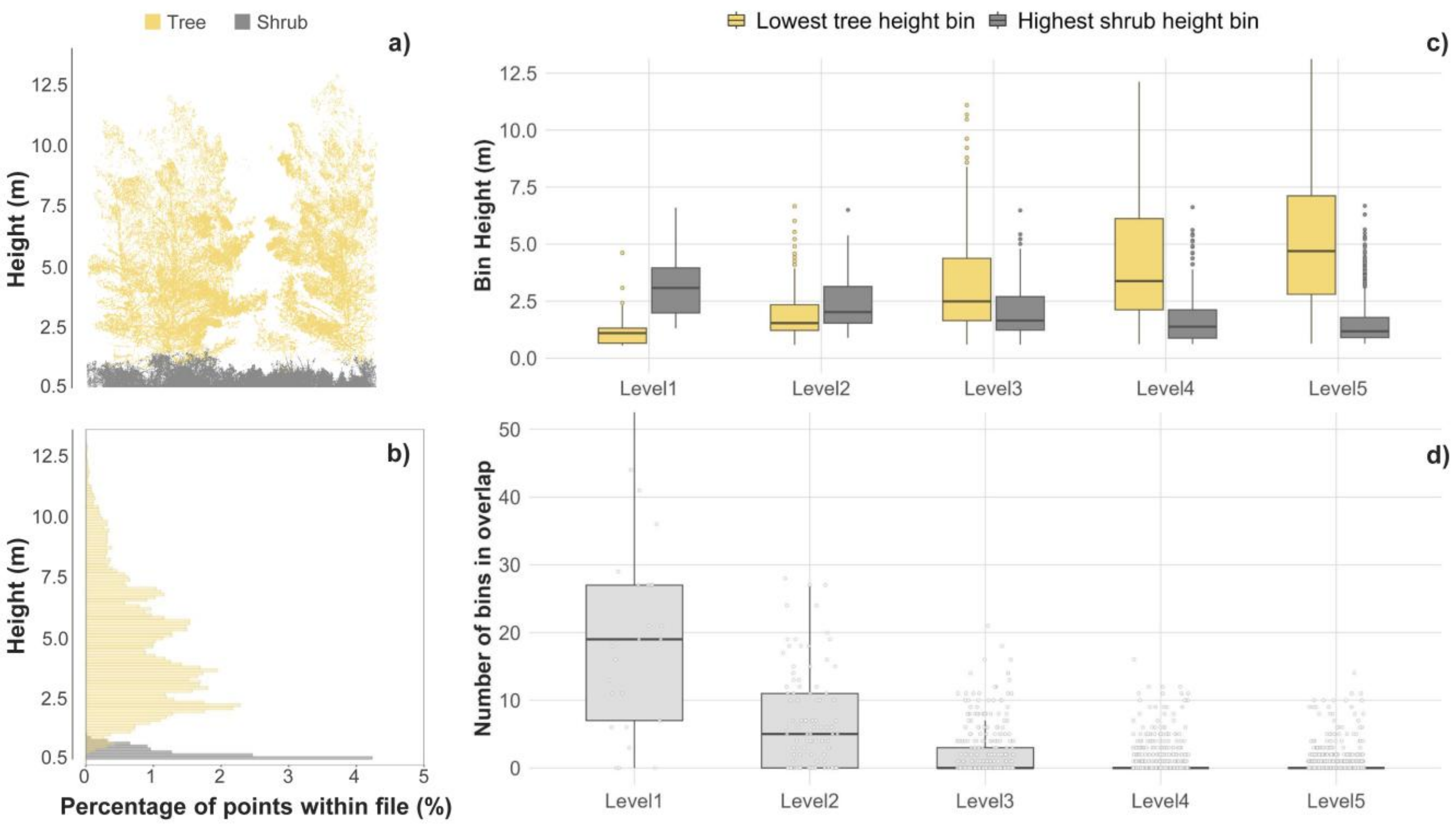

**Figure 5.** (a) Example of segmentation of the point cloud into shrub (gray) and tree (beige) strata. (b) Observed vertical distribution of points per bin and stratum for an example plot. (c) Representation of the bin corresponding to the minimum height of the tree stratum and the bin corresponding to the maximum height of the shrub stratum for each spatial unit and level. (d) Number of overlapping bins at each level.

Subsequently, each stratum was subdivided into bins (Figure 5b) following the same multiscale spatial scheme (Levels 1–5) described in Section 2.3.2 (Figure 4). Vertical discretization was performed using the same bin widths defined for each level in the Gaussian analysis, ensuring direct comparability between the observed and modeled forest structures. Accordingly, for each spatial unit, the observed vertical distribution of points per bin and per stratum (Figure 5c), and the number of overlapping bins per level (Figure 5d) were obtained.

From the segmented validation point cloud ($v$), for each spatial unit and height bin $zj$, the number of points in the shrub $N_{shrub}^{v}(z_j)$ and tree $N_{tree}^{v}(z_j)$ strata were computed. To ensure that the vertical continuity analysis was restricted to profiles exhibiting structural interaction between strata, only spatial units containing at least one vertical interval with simultaneous presence of both strata were considered. For bins satisfying this condition, the observed proportion of the tree stratum relative to the total was defined as:

$$P_{ts}(z_j) = \frac{N_{tree}^{v}(z_j)}{N_{tree}^{v}(z_j) + N_{shrub}^{v}(z_j)} \tag{4}$$

where $P_{ts}(z_j) \in [0,1]$ describes the relative dominance of the tree stratum at height bin $z_j$ along the vertical profile (Figure 6), with values greater than 0.5 indicating dominance of the tree stratum and values below 0.5 indicating dominance of the shrub stratum. $z_j$ denotes the vertical bin centered at a given height, $N_{shrub}^{v}(z_j)$ is the number of shrub points within $z_j$ and $N_{tree}^{v}(z_j)$ is the number of tree points within the same bin. This formulation provides a continuous representation of the observed vertical organization and serves as the empirical reference for validating the Gaussian model.

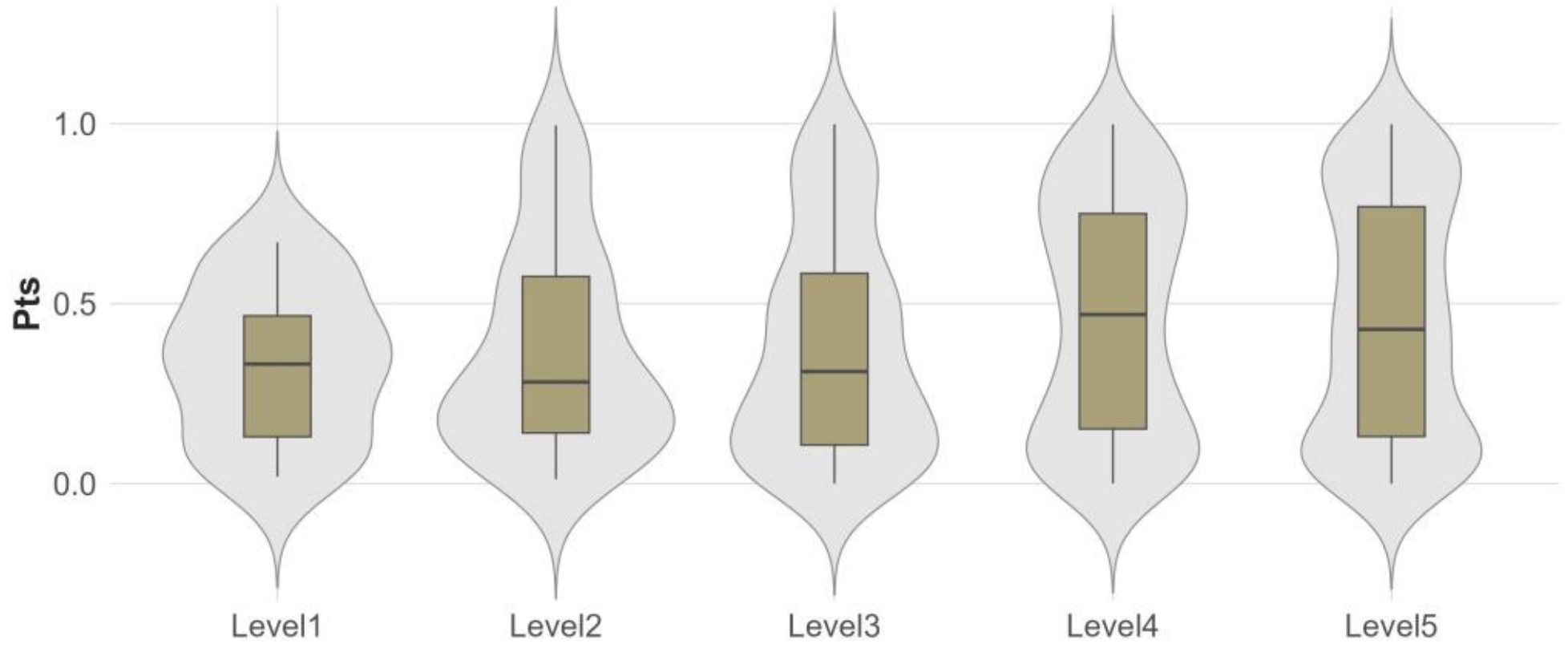

**Figure 6.** Representation of the relative dominance of the tree stratum ($P_{ts}$) at each level.

The study plots contained two main ecological fuel strata: shrub and tree, without an intermediate tree-regeneration layer. Based on the field-measured shrub heights and prior expert knowledge of the vegetation structure in the study area, 4 m was adopted as an operational upper threshold for assigning the lowest Gaussian component to the shrub stratum. Accordingly, when Gaussian 1 (G1) had $\mu_1 \leq 4$ m, it was assigned to the shrub stratum, and Gaussian 2 (G2), when present, represented the lower portion of the tree canopy. When G1 had $\mu_1 > 4$ m, the lowest detected component was considered to belong to the tree canopy; G1 was therefore reclassified as tree, G2 was omitted from the shrub-tree assignment, and the spatial unit was treated as lacking a shrub component. This assignment is based exclusively on the vertical ordering of the Gaussian components and does not imply a direct taxonomic identification of strata, but rather a geometric correspondence between the Gaussian components of the vertical profile and the strata defined in the reference segmentation. For complete G1-G2 pairs, curve 1 represents the shrub, curve 2 represents the lower portion of the tree canopy, while curve 3 typically corresponds to the mid-canopy, and curve 4, when present (Figure A2), is associated with the uppermost part of the canopy.

The ecological correspondence of this height-ordered assignment was evaluated post hoc for the spatial units containing a complete G1-G2 pair (Table A1). This assessment used the overlap-coefficient formulation reported for $OVL_{PDF}$ in Table 1, but applied here to compare each Gaussian component with each segmented reference stratum rather than to quantify the overlap between G1 and G2. For each spatial unit, the Gaussian mass of the component $i$, and the number of reference points belonging to stratum $s$, were normalized over the $j$ bins of the complete vertical profile:

$$\tilde{G}_i(z_j) = \frac{G_i(z_j)}{\sum_{k=1}^{J} G_i\,(z_k)} \tag{5}$$

$$\tilde{R}_s(z_j) = \frac{N_s^v(z_j)}{\sum_{k=1}^{J} N_s^v\,(z_k)} \tag{6}$$

where $i \in \{1,2\}$ represents G1 or G2, $s \in \{\text{shrub}, \text{tree}\}$ represents the segmented reference stratum, $G_i(z_j)$ is the Gaussian mass within bin $j$, and $N_s^v(z_j)$ is the number of reference points assigned to stratum $s$ within the same bin. The ecological correspondence between component $i$ and stratum $s$ was then calculated as:

$$OVL_{i,s}^{\text{ref}} = \sum_{j=1}^{J} \min\{\tilde{G}_i(z_j), \tilde{R}_s(z_j)\} \tag{7}$$

This coefficient ranges from 0, indicating no shared vertical distribution, to 1, indicating identical normalized vertical distributions. Because G1 and G2 constitute a single pair corresponding to two mutually exclusive reference strata, the two possible one-to-one assignments were compared jointly. The score of the height-ordered assignment was calculated as:

$$S_{\text{ordered}} = OVL_{1,\text{shrub}}^{ref} + OVL_{2,\text{tree}}^{ref} \tag{8}$$

whereas the score of the alternative assignment was calculated as:

$$S_{\text{swapped}} = OVL_{1,\text{tree}}^{ref} + OVL_{2,\text{shrub}}^{ref} \tag{9}$$

The height-ordered G1-shrub and G2-tree assignment was considered consistent with the segmented reference when $S_{\text{ordered}} > S_{\text{swapped}}$ and inconsistent with the segmented reference when $S_{\text{ordered}} < S_{\text{swapped}}$. Differences satisfying $|S_{\text{ordered}} - S_{\text{swapped}}| \leq 10^{-8}$ were classified as ambiguous. Spatial units lacking either a segmented reference stratum were classified as not evaluable because the two normalized reference distributions required for the comparison were not available. For descriptive interpretation of the inconsistent assignments, the four component-to-stratum overlap coefficients were also examined to determine whether the alternative score reflected greater overlap of G1 with the tree reference, greater overlap of G2 with the shrub reference, or both. This diagnostic examination did not modify the joint one-to-one classification. This reference-based assessment was calculated over the complete vertical profile and was used exclusively to evaluate the curve-to-stratum correspondence with the segmented reference; it was not used to reassign the Gaussian curves or to quantify shrub-canopy continuity.

For the subsequent reference-based validation of shrub-canopy continuity, the overlap interval between curves 1 and 2 was determined, and the bins within it were identified. This interval corresponds to the vertical zone where both components contribute simultaneously and represents the interaction between shrub and canopy fuels that is critical for the transition from surface to crown fires. For each Gaussian curve ($g$), and for each bin $zj$, the modeled contribution corresponding to the shrub stratum $N^{g}_{shrub}(z_j)$ and the tree stratum $N^{g}_{tree}(z_j)$ was computed. In addition, the partial area of each Gaussian within each overlapping bin was estimated, yielding $A_{shrub}(z_j)$ for the shrub stratum and $A_{tree}(z_j)$ for the tree stratum. For bins included within the overlap interval, the modeled proportion of the tree stratum was defined as:

$$A_{ts}(z_j) = \frac{A_{tree}(z_j)}{A_{tree}(z_j) + A_{shrub}(z_j)} \tag{10}$$

where $A_{ts}(z_j) \in [0,1]$ describes the modeled relative dominance of the tree stratum along the vertical profile, with values greater than 0.5 indicating dominance of the tree stratum and values below 0.5 indicating dominance of the shrub stratum.

To evaluate vertical coherence between the observed and modeled structures, the difference between the mean height of the observed overlap interval and the mean height of the Gaussian-derived overlap interval was computed for each stratum and level:

$$\Delta Z_{strata} = Z_{mean,observed} - Z_{mean,gaussian} \tag{11}$$

where $Z_{mean,observed}$ corresponds to the mean height (centroid) of the overlapping bins in the reference data, and $Z_{mean,gaussian}$ corresponds to the mean height (centroid) of the overlap modeled by the Gaussian curves.

Agreement between the observed and modeled vertical distributions was further assessed using Pearson correlation: (i) for the shrub stratum between $N^{g}_{shrub}(z_j)$ and $N^{v}_{shrub}(z_j)$; ii) for the tree stratum between $N^{g}_{tree}(z_j)$ and $N^{v}_{tree}(z_j)$; and iii) between $A_{ts}(z_j)$ and $P_{ts}(z_j)$.

In addition, the bin-wise error in the proportion of the tree stratum relative to the shrub stratum was estimated as the difference $P_{ts}(z_j) - A_{ts}(z_j)$, as well as its absolute value. Global coherence at the spatial unit level was evaluated using the WMSE, defined as:

$$WMSE = \frac{\sum_{j=1}^{N} w_j \left(P_{ts}(z_j) - A_{ts}(z_j)\right)^2}{\sum_{j=1}^{N} w_j} \quad (12)$$

where $w_j$ is the total number of observed points contributing to bin $j$ and $N$ is the number of bins under overlap that are common to both the reference-based and Gaussian-based representations. Consequently, bins supported by more points contributed more strongly to the error estimate. The weighted root mean squared error was calculated as $WRMSE = \sqrt{WMSE}$, which expresses the error in the original units of tree-stratum proportion. Both statistics were computed individually for each spatial unit and level. Level-specific uncertainty was represented by 95% percentile-based bootstrap confidence intervals for the median across spatial units, obtained from 2,000 resamples with replacement within each level.

## 3. Results

The results are presented in two sections. In Section 3.1 "Assessment of Gaussian decomposition", we evaluate the ability of the Gaussian decomposition methodology to reproduce the observed vertical structure of forest fuels relative to the segmented reference point cloud. Then, in Section 3.2 "Analysis of strata and fuel continuity metrics", we analyze the geometric, probabilistic, and overlap metrics for each analysis level.

### *3.1. Assessment of Gaussian decomposition*

The Gaussian decomposition was evaluated from two complementary perspectives: whether the height-ordered assignment of G1 to the shrub stratum and G2 to the tree stratum was consistent with the segmented reference distributions, and how faithfully the Gaussian decomposition reproduced the observed stratum profiles and their relative contributions within the shrub-tree transition, as assessed using Pearson correlations, centroid shifts, WMSE, and WRMSE.

#### 3.1.1 Consistency of the assignment of G1 to the shrub stratum and G2 to the tree stratum with the segmented reference

The Gaussian decomposition was applied across the multiscale gradient (Figure 4c), although the number of spatial units retained at successive processing stages, the availability of parameterizable profiles, and the number of detected Gaussian curves varied systematically with the spatial resolution of the analysis (Figure A2, Table A1).

Of the 3,829 available G1-G2 pairs, 3,182 contained both segmented reference strata and were therefore evaluable (Table 2). The height-ordered assignment was consistent with the segmented reference in 2,915 pairs (91.6% of the evaluable pairs) and inconsistent in 267 pairs (8.4%); no ambiguous scores were obtained. Consistency remained high across all analysis levels, ranging from 81.2% at Level 2 to 95.8% at Level 1. The remaining 647 complete pairs were not evaluable because either the shrub or tree reference stratum was absent from the corresponding spatial unit. The proportion of non-evaluable pairs increased as analysis levels became finer, reaching 21.6% at Level 5. Among the 267 pairs for which the height-ordered assignment was inconsistent with the segmented reference, 159 showed greater overlap of G2 with the shrub reference than with the tree reference, whereas 108 showed greater overlap of G1 with the tree reference than with the shrub reference.

**Table 2.** Evaluation of the consistency of the height-ordered assignment of G1 to the shrub stratum and G2 to the tree stratum with the segmented reference by analysis level. Available G1-G2 pairs correspond to spatial units containing both Gaussian components retained by the component-selection procedure (Table A1). Evaluable pairs additionally contained both shrub and tree distributions in the segmented reference. Assignment consistency was determined by comparing the joint score of the height-ordered assignment (Eq. 8) with that of the alternative one-to-one assignment (Eq. 9). Percentages of consistent, inconsistent, and ambiguous assignments were calculated relative to the number of evaluable pairs within each level. Pairs lacking either segmented reference stratum were classified as not evaluable.

| Level | Available G1-G2 pairs | Evaluable pairs | Consistent n (%) | Inconsistent n (%) | Ambiguous n (%) | Not evaluable n |
|---|---|---|---|---|---|---|
| 1 | 24 | 24 | 23 (95.8) | 1 (4.1) | 0 (0.0) | 0 |
| 2 | 80 | 80 | 65 (81.2) | 15 (18.8) | 0 (0.0) | 0 |
| 3 | 290 | 278 | 252 (90.6) | 26 (9.4) | 0 (0.0) | 12 |
| 4 | 847 | 771 | 716 (92.9) | 55 (7.1) | 0 (0.0) | 76 |
| 5 | 2,588 | 2,029 | 1,859 (91.6) | 170 (8.4) | 0 (0.0) | 559 |

3.1.2 Fit of Gaussian curves to the observed shrub and tree profiles

The correspondence between each Gaussian component and its reference stratum was evaluated using Pearson correlation between observed and modeled vertical point counts for the shrub and tree strata separately (Figure 7). For the shrub stratum, $N^{g}_{shrub}(z_j)$ and $N^{v}_{shrub}(z_j)$ were compared, whereas for the tree stratum, $N^{g}_{tree}(z_j)$ and $N^{v}_{tree}(z_j)$ were compared. At Level 1, correlations were generally high for both strata, whereas at Level 2, this pattern was maintained for the shrub stratum but became substantially weaker and more variable for the tree stratum (Figure 7a). From Level 2 onward, particularly for the tree stratum, an increase in the dispersion of correlation values is observed. For the shrub stratum, median correlations remained high across all levels, although dispersion and the number of outliers increased toward the finer analysis levels. At the same time, the median correlations remained positive, and a wider range of values appeared, reflecting the increasing structural heterogeneity among smaller spatial units. This trend became more pronounced at the finest spatial resolutions (Level 4 and Level 5), where some spatial units showed low or even negative correlations. These cases were mainly associated with profiles containing very few points or abrupt structural discontinuities. At these fine levels, the reduced size of the analysis units frequently generated partial or weakly structured vertical profiles, where local accumulations of a few points could dominate the Gaussian fitting (Figure 4c). Under these conditions, the assignment of Gaussian curves to the shrub and tree strata became less consistent, especially when the lower Gaussian component did not strictly correspond to the shrub layer.

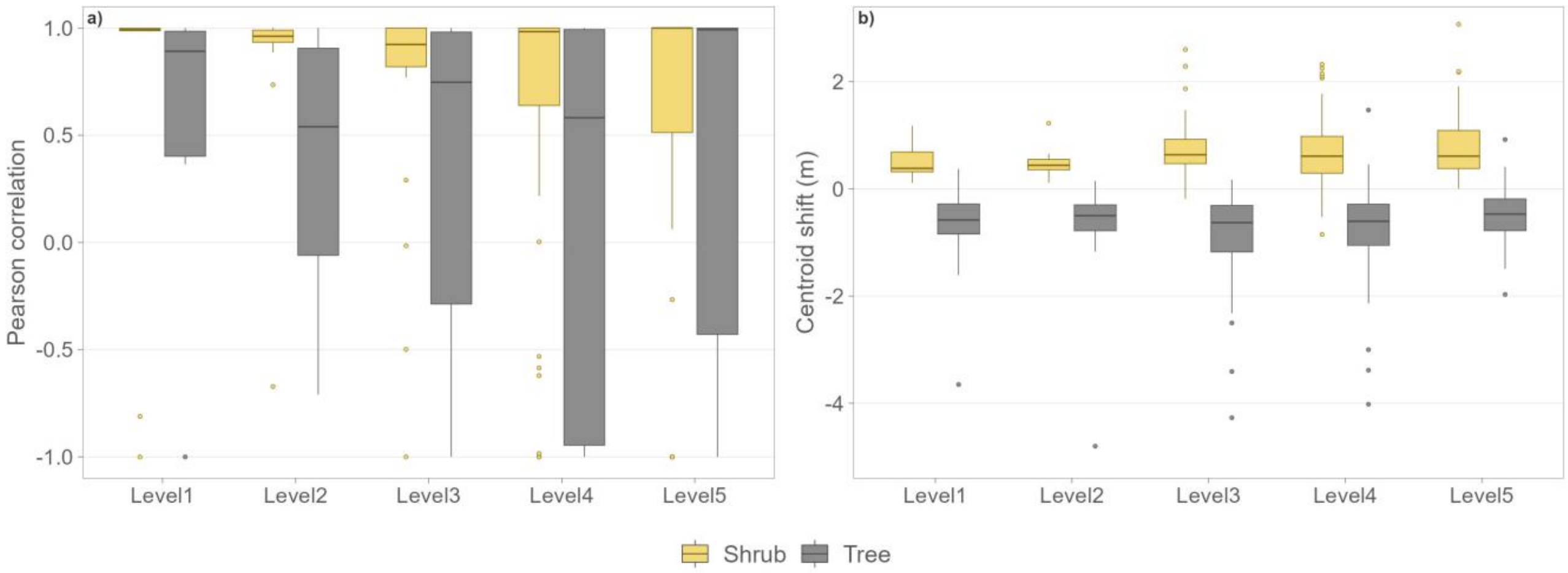

**Figure 7.** Comparison of shrub and tree strata (Gaussian pair 1–2) across the five analysis levels. (a) Pearson correlation between observed and Gaussian-derived numbers of points for the shrub $\left(N^{g}_{shrub}(z_j),\ N^{v}_{shrub}(z_j)\right)$ and tree strata $\left(N^{g}_{tree}(z_j), N^{v}_{tree}(z_j)\right)$. (b) Vertical centroid shift (Eq. 11) between observed and Gaussian-derived count distributions in the shrub and tree strata. Boxplots summarize the distribution of spatial-unit-level values separately by level, with shrub shown in yellow and tree in grey. Positive values indicate that the observed centroid is located above the Gaussian-derived centroid, whereas negative values indicate that the observed centroid is located below it.

The vertical position of the strata was also evaluated by comparing the shifts between the centroids of the observed vertical distributions and those derived from Gaussian decomposition (Figure 7b). The shift was calculated using Equation 11; positive values indicate that the observed centroid is located above the Gaussian-derived centroid, whereas negative values indicate that the observed centroid is located below it. For both strata, centroid shifts are generally relatively close to zero across the five analysis levels, although they showed a consistent directional difference between strata. The magnitude and dispersion of the shifts varied among analysis levels. However, a systematic pattern was observed between strata. For the shrub stratum, positive centroid shifts predominated, indicating that the Gaussian curve tended to underestimate the vertical position of the stratum and to be located slightly below the observed mean position of the shrub layer. This trend may be related to the accumulation of points in the lower part of the shrub layer, which shifts the centroid of the Gaussian distribution downward. By contrast, for the tree stratum, negative centroid shifts predominated, indicating that the Gaussian curve tended to overestimate the vertical position of the stratum and suggesting that the centroid derived from the tree Gaussian tended to lie slightly above the observed centroid. In this case, the observed points probably presented a distribution that was more extended downward, especially due to low branches, internal gaps, or partial scanner penetration, whereas the Gaussian curve tended to concentrate around the zone of maximum point density in the lower part of the canopy, generally located above the observed mean. The negative tail associated with the tree stratum became more evident at the finer spatial levels, where several pronounced negative outliers were observed.

The distribution of bin-level errors in the shrub-tree proportion was examined using the difference between observed and Gaussian-derived values for each vertical bin $\left(P_{ts}(z_j) - A_{ts}(z_j)\right)$ (Figure 8). At all spatial levels, error distributions were concentrated around values close to zero but shifted toward negative values, with negative median errors across all five analysis levels. This systematic shift toward negative values indicated that the tree proportion tended to be overestimated by the Gaussian representation. This tendency was already visible at Level 1 and was generally maintained across the multiscale gradient. The width of the distributions increased markedly from Level 1 to the intermediate analysis levels, but did not show a progressive increase toward the finest spatial scale. At Level 1, errors were concentrated within a relatively narrow range mostly below zero, whereas at Level 2, the distribution became wider. From Level 3 onward, the distributions remained broad, with both negative and positive errors becoming more frequent. At the finest levels (Level 4 and Level 5), the distributions continued to show substantial variability, although Level 5 presented a median closer to zero than the preceding levels. Extreme values in tree proportion error, close to ±1, corresponded mainly to bins with very few points, where small differences between observed and modeled proportions could generate high relative errors.

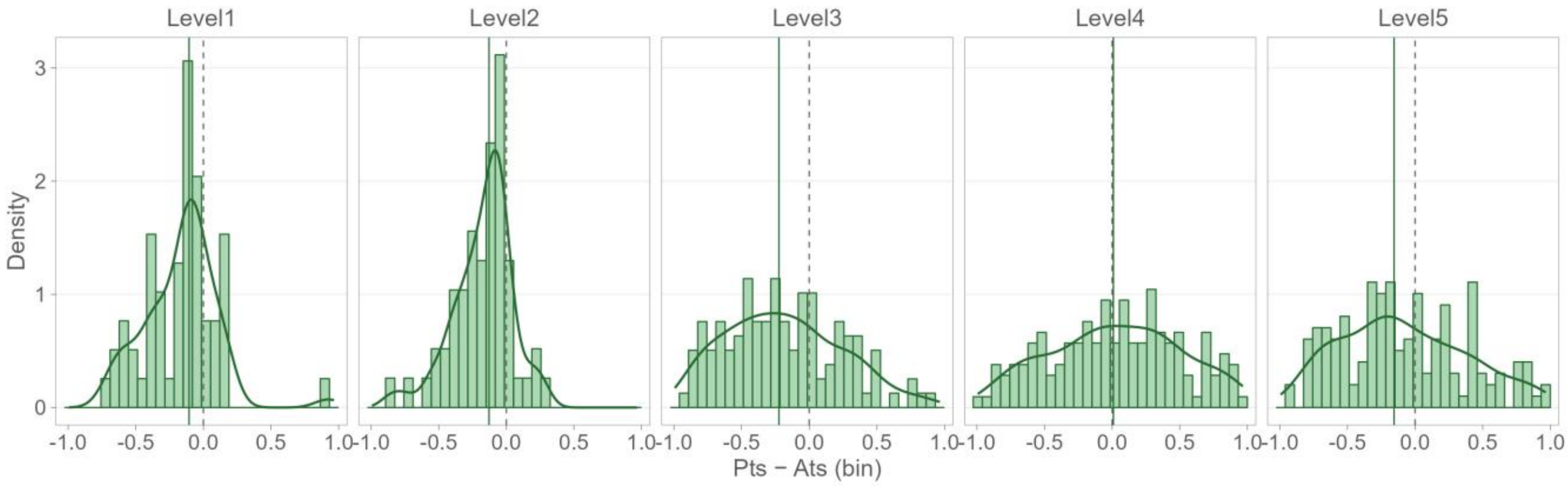

**Figure 8.** Distribution of bin-level errors in tree proportion $\left(P_{ts}(z_j) - A_{ts}(z_j)\right)$ across the five analysis levels. Histograms represent the empirical probability density of the errors using a common binning scheme

(30 bins) across all levels. The green curve shows the kernel density estimate (KDE), providing a smoothed representation of the distribution. The dashed vertical line marks zero error, facilitating visual assessment of bias, while the solid green vertical line indicates the median error for each level. All panels share the same x-axis range to allow direct comparison among levels.

Agreement between the observed tree-stratum proportion $P_{ts}(z_j)$ and the proportion modeled by the Gaussian decomposition $A_{ts}(z_j)$ was evaluated using profile-level Pearson correlation, the WMSE and the WRMSE computed for each spatial unit (Figure 9). Pearson correlation showed high values at most analysis levels (Figure 9a). At Level 1, the median correlation was 0.96 [95% CI: 0.93–1.00] and the interquartile range was very narrow, indicating that the Gaussian representation closely reproduced the overall shape of the observed profile at the plot scale. At Level 2, the median remained high (0.96 [0.89–0.99]), although the distribution widened and some isolated negative values appeared, reflecting greater variability among spatial units. At Level 3, the median again remained high (0.95 [0.86–1.00]), but the distribution showed greater width and a longer tail toward negative values. At Level 4, the median remained high (0.94 [0.76–1.00]) and dispersion decreased slightly relative to Level 3, although a larger number of negative outliers were observed. At Level 5, the distribution widened considerably and the median decreased to 0.83 [0.22–0.98], with a broad range of correlation values including numerous negative values, even though the median remained relatively high. The distribution of correlation values across predefined correlation classes is shown in Figure A3, which illustrates the increasing variability of correlation values at finer spatial scales and the particularly broad distribution observed at Level 5.

The overall performance of the methodology is summarized by the WMSE and WRMSE. WMSE was calculated from the bin-wise differences between observed and modeled proportions for each spatial unit and weighted by the number of points contributing to each bin (Eq. 12), whereas WRMSE was calculated as its square root (Figure 9b, c). Both metrics showed a scale-dependent, non-monotonic pattern. Errors increased from Level 1 to Level 3, where both WMSE and WRMSE reached their highest median values, and subsequently decreased at Levels 4 and 5. Variability among spatial units also increased from Level 1 and remained broad at the intermediate and finer analysis levels. This behavior was consistent with the bin-level error distributions shown in Figure 8, where the variability of discrepancies increased at intermediate analysis scales.

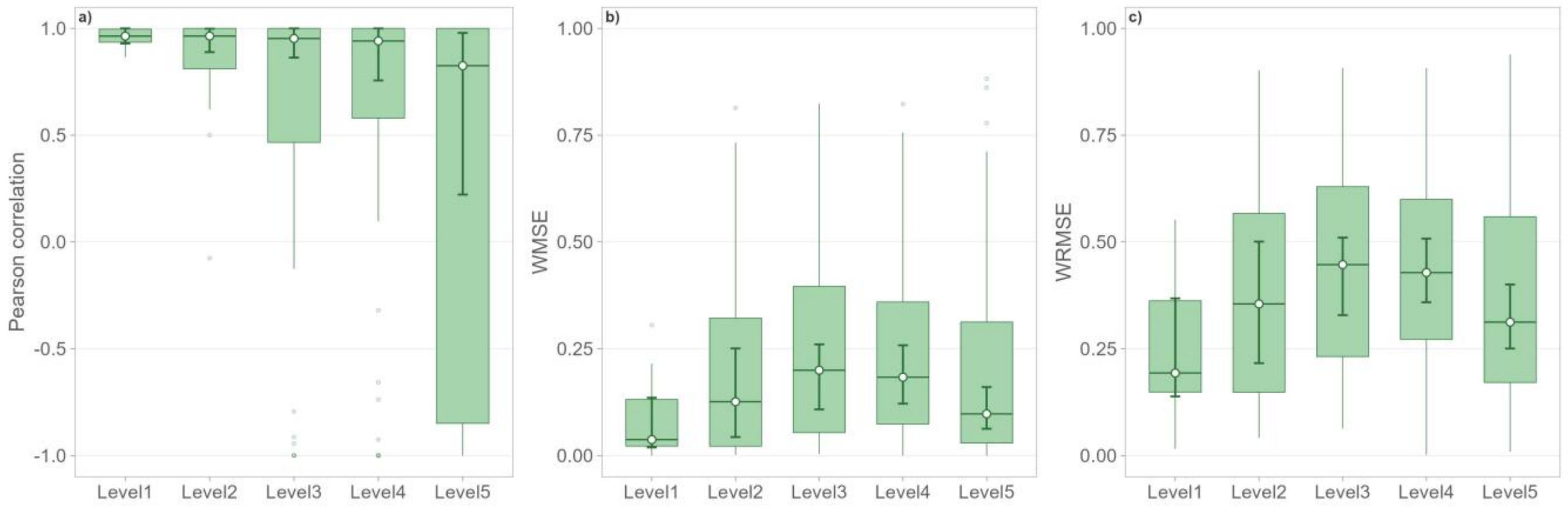

**Figure 9.** Validation metrics for the Gaussian-based estimates of the tree-stratum proportion across the five analysis levels. a) Pearson correlation between observed and Gaussian-derived tree-proportion values. b) Weighted mean squared error (WMSE) computed from the bin-wise differences between observed and Gaussian-derived tree-proportion values ($P_{ts}(z_j) - A_{ts}(z_j)$) across the bins common to both representations, using the number of points in each bin as weights; and (c) weighted root mean squared error (WRMSE), calculated as the square root of WMSE. Boxes represent the interquartile range, horizontal lines indicate the median, and whiskers extend to 1.5 times the interquartile range. Points and vertical bars show the median across spatial units and its 95% percentile-bootstrap confidence interval, based on 2,000 resamples and calculated separately for each level. Higher Pearson correlation indicates stronger agreement between observed and modeled proportions, whereas lower WMSE and WRMSE indicate lower estimation error.

In particular, at Level 1, error values were low, with a median WMSE of 0.04 [95% CI: 0.02–0.14] and WRMSE of 0.19 [0.14–0.37]. This indicated high agreement between the observed proportions and those derived from Gaussian decomposition when the analysis was performed at the plot scale. At Level 2, median error increased to 0.13 [0.04–0.25] for WMSE and 0.36 [0.22–0.50] for WRMSE, together with a substantial increase in dispersion among spatial units. At Level 3, both metrics reached their highest median values, with a WMSE of 0.20 [0.11–0.26] and a WRMSE of 0.45 [0.33–0.51]. This level also showed broad variability among spatial units, reflecting more pronounced differences between observed and modeled proportions along the vertical profile. At Level 4, median errors decreased slightly to 0.18 [0.12–0.26] for WMSE and 0.43 [0.36–0.51] for WRMSE, while dispersion remained high. Finally, at Level 5, median errors decreased further to 0.10 [0.06–0.16] and 0.31 [0.25–0.40] for WMSE and WRMSE, respectively, although the range of spatial-unit-level values remained broad.

### *3.2. Analysis of strata and fuel continuity metrics*

To illustrate how Gaussian decomposition captures different vertical structural configurations, Figure 10 presents three generic examples at the plot level (Level 1): (a) complete overlap between shrub and tree strata, (b) partial overlap with shrub and low branches, and (c) pronounced vertical discontinuity. For clarity, only a subset of the evaluation metrics is displayed in the figure.

In the case of the continuous structure (Figure 10a), the Gaussian components show a high degree of vertical coincidence, reflected in a low separation between means (|ΔH| = 1.57 m) and relatively low separability (d’ = 3.65; JM = 1.01). The overlap metrics confirm this behavior, with non-zero values ($OVL_{PDF}$ = 0.050; $OVL_{tree}$ = 0.105), indicating that both strata share a common vertical range. The intersection between the curves occurs at low heights, consistent with the proximity between the shrub distribution and the base of the canopy. In the example of partial overlap with low branches (Figure 10b), the separation between strata was considerably greater (|ΔH| = 6.76 m), and the separability metrics reached higher values (d’ = 4.64; JM = 1.22), indicating a clearer distinction between the Gaussian components. The overlap was minimal ($OVL_{PDF}$ = 0.009; $OVL_{tree}$ = 0.008), reflecting a limited vertical interaction between strata. However, the presence of low branches introduced intermediate elements that partially connected both layers, which was reflected in the shape of the distributions and in the persistence of a small but non-zero overlap. In the discontinuous structure (Figure 10c), the Gaussian components were clearly separated, with a large vertical distance between their means (|ΔH| = 7.64 m) and very high separability (d’ = 9.76; JM = 1.41). The overlap metrics were null ($OVL_{PDF}$ = 0.000; $OVL_{tree}$ = 0.000), indicating the absence of any shared vertical interval between strata. This configuration was characterized by a clear vertical gap between the shrub and canopy layers, reflecting a lack of structural connectivity.

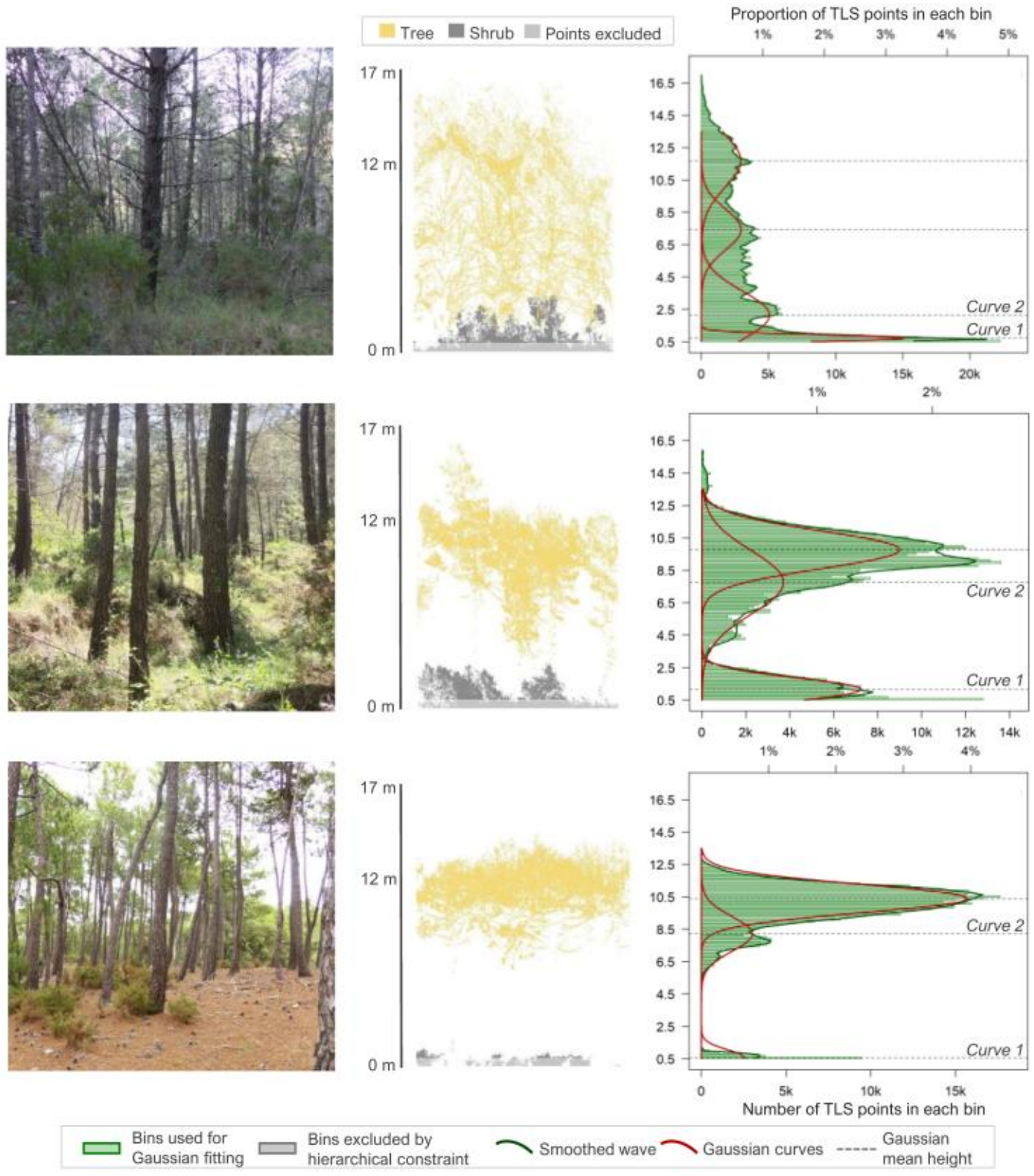

a)

| | A ± σ | μ ± σ | \|Δ\|H | d’ | JM | OVL [PDF] | OVL [tree] |
|---|---|---|---|---|---|---|---|
| Curve 1 | 26013 ± 871 | 0.79 ± 0.01 | 1.57 | 3.65 | 1.01 | 0.050 | 0.105 |
| Curve 2 | 6006 ± 650 | 2.36 ± 0.05 | | | | | |

b)

| | A ± σ | μ ± σ | \|Δ\|H | d’ | JM | OVL [PDF] | OVL [tree] |
|---|---|---|---|---|---|---|---|
| Curve 1 | 13833 ± 323 | 1.16 ± 0.02 | 6.76 | 4.64 | 1.22 | 0.009 | 0.008 |
| Curve 2 | 7863 ± 1474 | 7.92 ± 0.52 | | | | | |

c)

| | A ± σ | μ ± σ | \|Δ\|H | d’ | JM | OVL [PDF] | OVL [tree] |
|---|---|---|---|---|---|---|---|
| Curve 1 | 6405 ± 576 | 0.55 ± 0.11 | 7.64 | 9.76 | 1.41 | 0.000 | 0.000 |
| Curve 2 | 7561 ± 410 | 8.19 ± 0.16 | | | | | |

**Figure 10.** Examples of vertical structural configurations at the plot level (Level 1) derived from Gaussian decomposition. Three contrasting situations are shown: (**a**) complete overlap between shrub and tree strata,

(**b**) partial overlap, and (**c**) marked vertical discontinuity. These examples are illustrative and do not constitute analytical categories.

Across all valid spatial units, vertical separation and overlap between the shrub and tree strata exhibited scale-dependent patterns across the five analysis levels (Figure 11). These patterns were reflected in the geometric metrics (|ΔH|, Xcross1, Xcross2, and IoU (z = 2)), the probabilistic separability metrics (d', JM), and the overlap metrics ($OVL_{PDF}$ and $OVL_{tree}$).

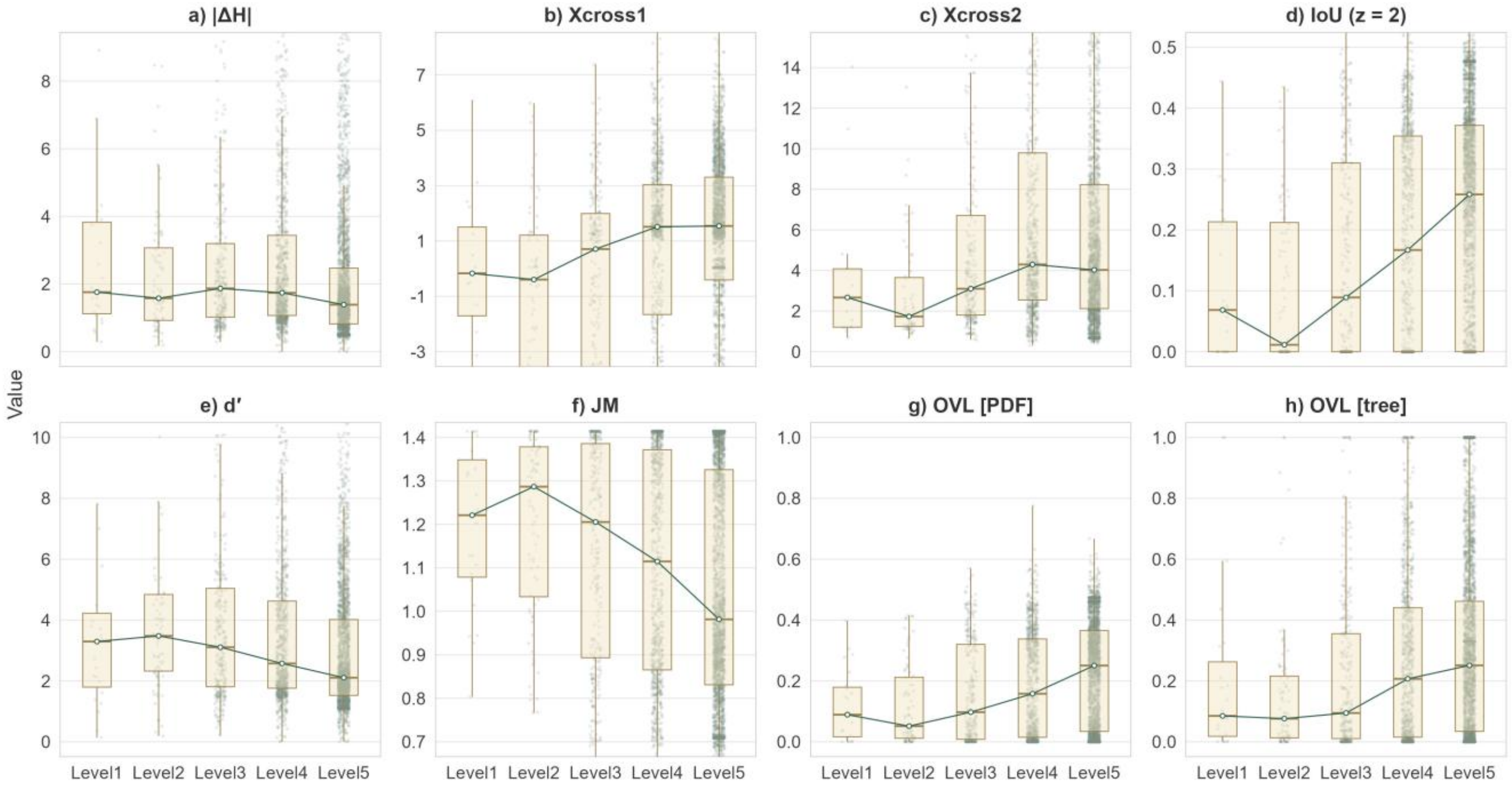

**Figure 11.** Multiscale behavior of structural descriptors derived from Gaussian decomposition for all valid shrub-tree Gaussian pairs. (a) Absolute height difference between Gaussian means (|ΔH|), (b) first intersection point (Xcross1), (c) second intersection point (Xcross2), (d) Intersection-over-Union for z = 2 (IoU), (e) d-prime distance (d′), (f) Jeffries–Matusita distance (JM), (g) probability density overlap ($OVL_{PDF}$), and (h) directional tree overlap ($OVL_{tree}$) across Levels 1–5. The dark green line connects the median values across levels, with white circles indicating the median at each level. Level 2 generally showed lower intersection and overlap values and greater separability, whereas from Level 2 toward the finer levels, intersection heights and overlap increased and separability decreased; |ΔH| remained comparatively stable through Level 4 and decreased at Level 5.

The absolute height difference between the Gaussian means (|ΔH|) remained relatively stable, with median values ranging from 1.58 to 1.87 m across the first four levels, before decreasing to 1.39 m at Level 5. The clearest scale-related change was therefore the lower |ΔH| at Level 5. This reduction at the finest level probably resulted in part from the progressive narrowing of the Gaussian-curve search range as the search progressed from coarse to fine levels (Eq. 3).

The intersection metrics (Xcross1 and Xcross2), which represent the heights at which the two Gaussian curves associated with the shrub and canopy strata intersect, showed a stronger dependence on the analysis level than |ΔH| (Figure 11). At the coarser levels (Level 1–Level 2), the Gaussian curves aggregated vegetation over larger spatial units, resulting in greater amplitude and standard deviation. As the level of detail increased (Level 3–Level 5), the search range of the Gaussian curves was reduced, the curves became narrower, and they reflected better-defined local configurations of the vertical structure, which was accompanied by a general upward displacement of the intersection points. Both intersection metrics had lower median values at Level 2 than at Level 1 and subsequently increased toward the finer analysis levels. Median Xcross1 increased from −0.40 m at Level 2 to 1.54 m at Level 5, whereas median Xcross2 increased from 1.74 m at Level 2 to a maximum of 4.30 m at Level 4, followed by a slight decrease to 4.03 m at Level 5. Both intersection metrics also exhibited broad distributions, particularly at the finer analysis levels. This behavior reflected the combined effects of spatial scale and fitting constraints: finer units reduced the search range and captured more localized profiles, while the intersection points remained mathematically sensitive to differences in curve width and tail behavior. Consequently, their heights should not, by themselves, be interpreted as direct evidence of structural proximity between the shrub and canopy strata.

Regarding the statistical separability metrics (Figure 11), the d' index increased slightly from Level 1 to its highest median at Level 2 (3.48), then decreased to 2.11 at Level 5. The JM distance behaved similarly, reaching its highest median at Level 2 (1.29) before decreasing progressively to 0.98 at Level 5, although these final values indicate that appreciable separability was generally retained. Conversely, median IoU (z = 2) decreased from 0.069 at Level 1 to its minimum of 0.011 at Level 2 and then increased progressively to 0.258 at Level 5. Thus, Level 2 was characterized by greater modeled separability and lower parametric-interval overlap, whereas from Level 2 toward the finer levels, separability decreased and interval overlap increased. Because fit assessment also showed substantial variability among spatials units at intermediate and finer analysis levels (Section 3.1.2), these trends may reflect both localized structural configurations and greater fitting sensitivity to sparse or partial profiles, rather than unequivocally greater physical vertical continuity.

Considering the overlap metrics between G1 (shrub stratum) and G2 (tree stratum), Figure 11 shows the distribution of $OVL_{PDF}$ and $OVL_{tree}$ for all valid spatial units across Levels 1–5. Both metrics decreased slightly from Level 1 to Level 2 and subsequently increased toward the finer levels. Median $OVL_{PDF}$ decreased from 0.089 at Level 1 to 0.052 at Level 2 and then increased progressively to 0.251 at Level 5, whereas median $OVL_{tree}$ decreased from 0.085 to 0.076 between Levels 1 and 2 before increasing to 0.251 at Level 5. The broad distributions, extending from values close to zero to high overlap, reflected the combination of spatial units with diverse forest structures, ranging from clearly separated strata to substantial shared vertical space. At Level 1, the plot-level variation in $OVL_{tree}$ and its correspondence with the TLS-derived shrub-tree structure are illustrated in Figure A4. Therefore, the pooled medians do not imply a general absence of vertical continuity; rather, the range of values captures the structural heterogeneity present within each level, as illustrated by the contrasting examples in Figure 10. At Levels 4 and 5, however, individual high overlap values should be interpreted cautiously because the substantial variability in fit performance observed at finer levels (Section 3.1.2) may increase the influence of Gaussian-tail overlap in sparse or partial profiles.

## 4. Discussion

In this study, we developed and evaluated a methodology based on the Gaussian decomposition of vertical profiles derived from TLS point clouds across 25 Mediterranean forest plots to characterize vertical fuel continuity. This approach enabled a consistent representation of the vertical interaction between shrub and tree strata, capturing a wide range of structural configurations, from complete overlap to full discontinuity (Figure 10), while revealing broad scale-dependent distributions across all valid spatial units (Figure 11), and providing explicit quantitative metrics of vertical connectivity. This capability is particularly relevant in Mediterranean ecosystems, where the coexistence of shrub and tree strata promotes vertical fuel connectivity and strongly influences fire propagation dynamics (Hakkenberg et al., 2024; Rodríguez y Silva et al., 2017). The main contribution of this work lies in applying Gaussian decomposition, traditionally applied to LiDAR signal processing or structural layer identification (Zhou et al., 2017; Zhuang et al., 2015), to vertical profiles derived from discrete-return TLS point clouds and in proposing and analyzing derived fuel continuity metrics that could serve as a more complete characterization of the understory and crown vegetation, with potential relevance to fire behavior and wildfire risk analysis. Within this framework, each Gaussian curve is associated with a functional vegetation stratum, and their overlap is interpreted as a quantitative indicator of vertical continuity. This approach complements the discrete fuel types with a continuous and interpretable representation of fuel vertical structure. The results show that the high point density provided by TLS enables a clear identification of transitions between strata at the coarser analysis levels, particularly at Level 1, where the vertical profiles are more completely sampled (Figures 7–9). However, as the scale of analysis decreases and spatial coherence in point distribution is reduced (Levels 4 and 5), the representational capacity becomes more variable, particularly in terms of profile correlation and stratum-specific fit (Figures 7–9).

### *4.1. Performance of Gaussian decomposition*

The performance of the Gaussian decomposition methodology is conditioned by the scale of analysis and by the quality of the structural information contained in the vertical profiles (Figure 4c). In this regard, the results show a progressive loss of valid profiles at finer scales (Figure A2 and Table A1), where structural gaps become increasingly evident. This behavior is associated with the reduction of the analysis area and the explicit emergence of structural discontinuities, such as canopy openings, local absence of the shrub layer, or the simultaneous absence of both strata, resulting in vertical profiles with insufficient information for multimodal parameterization. This pattern is consistent with previous studies showing that forest structure characterization using LiDAR is sensitive to point density and processing methodology. For instance, Viedma and Moreno (2025) highlight that estimates of ladder-fuel structural metrics degrade under conditions of low point density or high structural heterogeneity. Similarly, the literature on Gaussian decomposition of full-waveform LiDAR signals has reported recurrent limitations related to the detection of weak components, the merging of overlapping peaks, and the emergence of spurious components associated with noise (Liu et al., 2025; Mountrakis and Li, 2017). In addition, the configuration of the Savitzky–Golay filter introduces a trade-off in the detection of Gaussian curves, where excessive smoothing (large window sizes) tends to oversimplify the signal, merging or removing nearby peaks, while insufficient smoothing may generate spurious maxima, leading to an overestimation of the number of curves (Savitzky and Golay, 1964). This effect is more pronounced at finer scales, where lower point density limits curve detectability. This behavior is analogous to ALS signal processing, where decomposition has been shown to be sensitive to preprocessing choices and fitting parameterization, particularly in complex signals or those with partially overlapping components (Mountrakis and Li, 2017; Zhou et al., 2017). In our case, the adopted parameterization (fixed polynomial order $p = 2$ and scale-adapted window sizes) represents a compromise that preserves the global shape of the profile while limiting artificial variability. These results show that the interpretability of vertical profiles depends directly on the structural coherence of the point cloud and the spatial scale of analysis.

The interpretation of the metrics (Figure 11) should focus on their scale-dependent behavior across diverse configurations of shrub-tree interaction. In this study, the behavior of $|\Delta H|$ reflected the vertical separation between the shrub and tree strata, remaining relatively stable from Levels 1 to 4 and decreasing at Level 5 in both years. This pattern suggests that the distance between the means of the Gaussian curves ($\mu$) is relatively stable across most spatial resolutions, with a clearer scale effect emerging at Level 5. In contrast, the changes observed in Xcross1 reflect an upward shift in the relative position of the intersection between Gaussian curves as the level of analysis becomes finer. Changes in intersection metrics (Xcross1 and Xcross2) are primarily associated with the effect of scale on the shape and width of the Gaussian curves, as well as with the range-bounding methodology used in Gaussian searches, rather than with fundamental changes in the vertical relationship between strata. In this sense, these metrics should be interpreted as descriptors of the relative positioning of the distributions, while $|\Delta H|$ and the separability metrics provide complementary descriptions of their separation. The general decrease in separability metrics (d′ and JM) from Level 2 onward, accompanied by increasing IoU, indicates lower modeled separability and greater parametric-interval overlap toward the finer levels. Because the pooled distributions combine heterogeneous forest configurations, these scale-dependent trends should not be interpreted as a uniform shift toward greater physical vertical continuity, particularly at Levels 4 and 5, where fit performance was more variable.

The behavior of the Gaussian decomposition must be interpreted in relation to the underlying forest structural configurations illustrated in Figure 10. In simple vertical structures (Figure 10c), characterized by a single dominant canopy stratum and limited understory development, Gaussian decomposition yields stable and well-defined curves, allowing structural parameters to be directly estimated. These results are consistent with those reported by Martin-Ducup et al. (2025) who showed that ALS-derived structural metrics perform better in homogeneous stands with clear stratification. In two-layer structures (Figure 10b), where a shrub layer is clearly separated from the tree canopy, the decomposition remains robust, although the relative position of the Gaussian components becomes more sensitive to local variations in point distribution. These configurations correspond to classic ladder fuel scenarios, in which vertical fuel continuity is partial but structurally significant (Marino et al., 2018). In contrast, multilayered and vertically continuous structures (Figure 10a) represent the most complex configurations for characterizing the vertical organization of fuels. In these cases, the vertical distribution of points does not necessarily exhibit clearly isolated modes, and substantial overlap among vegetation elements is expected. However, unlike previous ALS-based approaches that report difficulties in resolving such complex arrangements due to occlusion effects and structural mixing (Kacic et al., 2025; Marino et al., 2018; Martin-Ducup et al., 2025; Viedma and Moreno, 2025) the Gaussian decomposition applied in this study remains capable of identifying and separating the main structural components. This indicates that, even under conditions of high vertical continuity, the signal retains sufficient structure to be decomposed into meaningful Gaussian elements. Rather than failing under complex scenarios, the method captures the gradual transition from distinct strata to increasingly overlapping ones, enabling a more explicit representation of vertical fuel connectivity. This TLS capability is particularly relevant in multilayered forests, where traditional ALS metrics (e.g., CBH, FSG) or voxelization may be insufficient to describe the vertical arrangement of fuels.

Within this structural context, overlap-based indicators provide a structurally meaningful proxy for vertical continuity between shrub and tree strata (Figure 11g and 11h). Low overlap values are associated with structurally separated configurations (Figure 10c), where distinct vertical strata can be clearly identified. In contrast, higher overlap values occur in complex and vertically continuous structures (Figure 10a and 10b), where the distinction between strata progressively blurs and both Gaussian curves share a common vertical space. It is important to note that the observed variability in overlap reflects both the diversity of structural configurations represented within each level and the increase in structural heterogeneity of the point cloud toward finer scales of analysis. At coarser scales (Levels 1 and 2), point aggregation produces smoother, more stable vertical profiles, enabling more consistent identification of dominant structural components and, consequently, more reliable overlap estimates. In contrast, finer spatial subdivisions increase point cloud fragmentation, amplifying local variability and reducing the robustness of the decomposition. In this sense, the results suggest that, for the generation of quantitative vertical fuel-continuity TLS-derived maps, a spatial resolution between approximately 3 and 6 m (Levels 2 and 3) may provide a useful compromise between spatial detail and robustness in the characterization of shrub-canopy connectivity. Such maps, together with other fire risk variables, could help identify areas where preventive forest management may be prioritized. For operational screening, $OVL_{tree}$ could be mapped by its distribution within each management area, with higher values indicating greater shrub-tree overlap and identifying candidate areas for field assessment or the prioritization of preventive treatments such as understory clearing or pruning. These decisions should be taken together with other structural variables, including stand density, horizontal fuel continuity, and fragmentation. Since we only have TLS data in separate plots and cannot create such maps, Figure A4 shows instead an example of the point clouds per plot colored attending to the corresponding values of $OVL_{tree}$ for level 1.

The reference-based one-to-one comparison showed that the assignment of G1 to the shrub stratum and G2 to the tree stratum was consistent with the segmented reference in 91.6% of the evaluable spatial-unit pairs (Table 2), indicating that vertical ordering generally agreed with the segmented reference for the first two Gaussian components. However, this result was conditional on the presence of both reference strata; 16.9% of the available pairs were not evaluable, particularly at the finer levels of analysis. The inconsistent assignments occurred mainly when G2 showed greater vertical overlap with the shrub reference or G1 with the tree reference, indicating that height ordering can become less reliable in fragmented or weakly structured profiles.

These scale-dependent limitations are also reflected in the reference-based profile-level validation, which shows that Gaussian decomposition maintains a strong ability to reproduce the transition between strata along the vertical profile, as indicated by the high Pearson correlation values across most levels (Figure 9). However, the increase in dispersion and the occurrence of negative correlations at finer scales indicate that this capability is not uniform but rather depends directly on the structural coherence of the profile. While the shrub stratum is consistently represented across all scales except Level 5, the tree stratum shows lower median correlations and substantially greater dispersion (Figure 7a). This behavior is consistent with the TLS sensor configuration relative to the ground and contrasts with the pattern typically observed in ALS data (Hancock et al., 2017). In addition, the structural complexity of Mediterranean canopies, characterized by irregular crowns and highly heterogeneous vertical distributions, leads to variable point densities along the TLS profile due to occlusion effects caused by branch arrangement. In this context, the bin-by-bin representation of the tree stratum is particularly sensitive to scale changes, due to the combined effects of occlusion, variability in point density, and reduced point cloud quality in the upper canopy layers. In addition, the behavior of WMSE and WRMSE confirms that estimation accuracy does not follow a monotonic trend with scale, and that reductions in error at finer levels do not necessarily imply a better representation of the transition between strata. Instead, they often reflect a simplification of vertical profiles within very small spatial units. This is particularly evident at Level 5, where relatively low median WMSE and WRMSE coexist with a lower median Pearson correlation and the greatest dispersion of correlation values, showing that error magnitude and agreement in profile shape capture different aspects of Gaussian performance. The combined analysis of correlation and error highlights that the shrub-tree transition is robustly captured at scales where the profile retains sufficient structural coherence, whereas at finer scales, fragmentation introduces variability in the representation without necessarily reflecting real changes in vertical continuity. At Levels 4 and 5, the reduced size of the analysis units frequently produces weakly structured or partial profiles, limiting the ecological interpretability of the Gaussian components and increasing variability in the representation of shrub-canopy relationships. These discrepancies do not arise from a systematic failure of the Gaussian model, but rather from the inherent difficulty of representing highly irregular or partially observed profiles using a limited number of Gaussian components.

*4.2. Limitations and future work*

A potential limitation of our validation process is that the manual segmentation of tree and shrub strata is inherently subject to human operator bias (Laino et al., 2026). To minimize this source of uncertainty, segmentation was conducted in a systematic manner and carefully supervised, and boundary assignment was supported by TLS-acquired imagery and field photographs, in addition to point-cloud attributes. In addition, no independent field validation of this segmentation was carried out, as the original field campaign was not designed for this purpose, which constitutes an additional constraint. However, obtaining such measurements in the field would likely have been less accurate than direct measurements derived from the point cloud itself. The use of TLS as a high-resolution reference is consistent with studies that have used TLS as ground-truth validation for deriving full-waveform ALS metrics (Crespo-Peremarch et al., 2018; Hancock et al., 2017), for validating airborne laser scanning acquired from Unmanned Aircraft System (UAS) platforms (Puliti et al., 2020), or for generating manually labeled point clouds for training machine learning models (Laino et al., 2026). On the other hand, it is important to emphasize that, unlike other studies that incorporate radiometric information (e.g., LiDAR intensity) to improve discrimination between vegetation layers (Yépez-Rincón et al., 2021), this study deliberately adopts an approach based exclusively on point cloud geometry. This decision allows isolating the role of three-dimensional structure in characterizing vertical fuel continuity, avoiding additional dependencies related to radiometric calibration or acquisition conditions (Torralba et al., 2026).

A separate limitation concerns the ecological scope of the component-to-stratum interpretation. The study plots represented a limited set of Mediterranean forest configurations composed of a shrub layer and an overstory tree layer, without a distinct tree-regeneration stratum.

Consequently, the ecological interpretation of the ordered Gaussian components should not be generalized to forests containing regeneration or more complex understory stratification without additional validation. Taken together, these limitations and methodological choices reflect the inherent challenges of analyzing TLS point clouds in complex forest environments, while also highlighting the potential of geometric approaches as a transferable foundation for the development of future, more automated and generalizable methodologies.

Unlike conventional ALS-based metrics, this approach enables an explicit description of vertical fuel continuity, a key factor in fire propagation that has traditionally been simplified within FBFT frameworks (Aragoneses and Chuvieco, 2021). The proposed method not only allows the identification and differentiation of the main vegetation strata but also quantifies their interaction through explicit transition and overlap metrics, directly addressing one of the main gaps in traditional approaches. In this context, recent advances in mobile laser scanning (MLS) systems have demonstrated their potential for characterizing fuel structure at high spatial resolution over large areas. For instance, Hoffrén et al. (2024) showed that handheld mobile laser scanning can effectively capture the vertical distribution of fuels and support the definition of Prometheus fuel types in structurally complex Mediterranean forests. Building upon this type of structural information, the methodology proposed in this study could be extended to MLS datasets to move beyond discrete fuel-type classification and provide a continuous quantification of vertical fuel connectivity. However, the transferability of this approach to MLS and especially ALS data requires careful consideration of scale effects, as lower point densities and different acquisition geometries may limit the detection of fine vertical structures. In this regard, the multiscale framework presented here provides a suitable basis to explicitly evaluate how structural metrics behave under varying spatial resolutions, which is essential for adapting the methodology to large-area applications.

Vertical fuel continuity is a key factor influencing the transition from surface to crown fires and, therefore, fire propagation dynamics. These structural metrics could be used as complementary descriptors for forest fuel inventory, characterization and preventive management, particularly in forest systems with complex and heterogeneous structures, where conventional fuel typologies are often insufficient. In this regard, the results of this study open the door to integrating structural metrics derived directly from TLS point clouds into fuel assessment frameworks, supporting the identification of areas where vertical fuel connectivity may warrant field assessment or management intervention. However, the potential contribution of these metrics to fire behavior modeling and wildfire risk assessment was not evaluated in the present study. Future research should therefore test their relationships with observed or simulated fire-behavior variables before considering their potential use in fire-behavior modeling or risk assessment.

## 5. Conclusions

A method to characterize vertical fuel continuity based on the Gaussian decomposition of vertical profiles derived from TLS point clouds has been presented. The proposed approach characterizes the transition between shrub and tree strata and quantifies understory and canopy continuity through different metrics, providing continuous structural descriptors that could complement the traditional fuel types used in wildfire risk assessment. The results show that Gaussian decomposition can robustly reproduce the vertical interactions between strata at coarse and intermediate analysis scales, with the profiles retaining sufficient structural coherence. Under these conditions, the method allows for the consistent identification and quantification of vertical fuel continuity, providing an interpretable representation of vegetation vertical organization. In particular, overlap- and separability-based metrics derived from the Gaussian decomposition showed contrasting values among illustrative continuous, partially connected, and discontinuous shrub-canopy configurations. However, the performance of the method is clearly dependent on the scale of analysis. At finer scales, spatial fragmentation, reduced point density, and increased profile irregularity limit the stability of the decomposition, increasing metric variability and reducing the proportion of valid profiles. This finding highlights that higher spatial resolution does not necessarily lead to a better characterization of vertical structure, but rather that a trade-off exists between the level of detail and the structural coherence of the vertical profile. These results suggest that spatial resolutions between approximately 3 and 6 m (Levels 2–3) are the most suitable for quantitative mapping of TLS-derived vertical fuel continuity, providing an effective compromise between spatial detail and the consistent representation of shrub-canopy connectivity patterns. In addition, the results show that the evaluation of vertical continuity cannot rely solely on error metrics. Low error values (WMSE and WRMSE) may be associated with simplified representations of the vertical profile, particularly in spatial units with low point density, reinforcing the need to jointly interpret errors, correlation, and structural metrics.

Although this study was developed using TLS data, the proposed framework could potentially be extended to MLS and ALS datasets, where large-area characterization of fuel continuity is operationally more feasible. However, the lower point density and different acquisition geometries of these systems may affect the representation of fine vertical structures, making the analysis scale a critical factor for the transferability of the methodology. Nevertheless, this advantage is constrained to specific analysis scales and requires methodological strategies to control the effects of fragmentation and occlusion. We consider that the proposed approach represents a significant step forward toward the incorporation of continuous vertical connectivity metrics into FBFT. The ability to explicitly quantify interactions between strata could provide a basis for complementing existing fuel models, particularly in ecosystems characterized by high structural heterogeneity. Subsequent mapping of the described metrics could help identify areas most prone to vertical fire spread, supporting preventive forest management practices. Potential applications of these contributions also include improved characterization of ladder fuels and strengthening decision-making support in wildfire risk assessment.

**Author Contributions:** Conceptualization, J.T., L.Á. and P.C-P; methodology, J.T., L.Á. and P.C-P; software, J.T. and P.C-P; validation, J.T., L.Á., P.C-P and J.P.C-R; formal analysis, J.T., and L.Á.; investigation, J.T.; data acquisition, J.T., P.C-P, M.S-M and J.P.C-R; data curation, J.T.; writing—original draft preparation, J.T.; writing—review and editing, J.T., L.Á., P.C-P, M.S-M and J.P.C-R; visualization, J.T.; funding acquisition, L.Á. All authors have read and agreed to the published version of the manuscript.

**Funding:** The work presented in this publication was supported by grants PID2020-117808RB-C21 and PID2024-158591OB-I00, funded by MICIU/AEI/10.13039/501100011033 and the ESF+. Funding for open access charge: Universitat Politècnica de València (PAID-12-26).

**Declaration of competing interests:** The authors declare no conflicts of interest.

**Declaration of generative AI use:** The authors declared that Generative AI was not used in the creation of this manuscript.

## References

Agee, J.K., 1996. The Influence of Forest Structure on Fire Behavior, in: 17 Th Forest Vegetation Management Conference. pp. 52–68.

Andersen, H.E., McGaughey, R.J., Reutebuch, S.E., 2005. Estimating forest canopy fuel parameters using LIDAR data. Remote Sens. Environ. 94, 441–449. https://doi.org/10.1016/j.rse.2004.10.013

Aragoneses, E., Chuvieco, E., 2021. Generation and Mapping of Fuel Types for Fire Risk Assessment. Fire 4, 59. https://doi.org/10.3390/fire4030059

Atchley, A.L., Linn, R., Jonko, A., Hoffman, C., Hyman, J.D., Pimont, F., Sieg, C., Middleton, R.S., 2021. Effects of fuel spatial distribution on wildland fire behaviour. Int. J. Wildland Fire 30, 179–189. https://doi.org/10.1071/WF20096

Atkins, J.W., Bohrer, G., Fahey, R.T., Hardiman, B.S., Morin, T.H., Stovall, A.E.L., Zimmerman, N., Gough, C.M., 2018. Quantifying vegetation and canopy structural complexity from terrestrial Li <scp>DAR</scp> data using the <scp>forestr r</scp> package. Methods Ecol. Evol. 9, 2057–2066. https://doi.org/10.1111/2041-210X.13061

Bergen, K.M., Gilboy, A.M., Brown, D.G., 2007. Multi-dimensional vegetation structure in modeling avian habitat. Ecol. Inform. 2, 9–22. https://doi.org/10.1016/j.ecoinf.2007.01.001

Bhattacharyya, A., 1943. On a measure of divergence between two statistical populations defined by their probability distributions. Bulletin of the Calcutta Mathematical Society 35, 99–109.

Chergui, B., Fahd, S., Santos, X., Pausas, J.G., 2018. Socioeconomic Factors Drive Fire-Regime Variability in the Mediterranean Basin. Ecosystems 21, 619–628. https://doi.org/10.1007/s10021-017-0172-6

CloudCompare, 2024. CloudCompare (version 2.13.1) [GPL software].

Coverdale, T.C., Davies, A.B., 2023. Unravelling the relationship between plant diversity and vegetation structural complexity: A review and theoretical framework. Journal of Ecology 111, 1378–1395. https://doi.org/10.1111/1365-2745.14068

Crespo-Peremarch, P., Tompalski, P., Coops, N.C., Ruiz, L.Á., 2018. Characterizing understory vegetation in Mediterranean forests using full-waveform airborne laser scanning data. Remote Sens. Environ. 217, 400–413. https://doi.org/10.1016/j.rse.2018.08.033

Cruz, M.G., Alexander, M.E., Wakimoto, R.H., 2004. Modeling the Likelihood of Crown Fire Occurrence in Conifer Forest Stands. Forest Science 50, 640–658. https://doi.org/10.1093/forestscience/50.5.640

Domingo, D., de la Riva, J., Lamelas, M., García-Martín, A., Ibarra, P., Echeverría, M., Hoffrén, R., 2020. Fuel Type Classification Using Airborne Laser Scanning and Sentinel 2 Data in Mediterranean Forest Affected by Wildfires. Remote Sens. (Basel). 12, 3660. https://doi.org/10.3390/rs12213660

Elzhov, T. V., Mullen, K.M., Spiess, A.-N., Bolker, B., 2023. minpack.lm: R Interface to the Levenberg-Marquardt Nonlinear Least-Squares Algorithm Found in MINPACK, Plus Support for Bounds.

Erdody, T.L., Moskal, L.M., 2010. Fusion of LiDAR and imagery for estimating forest canopy fuels. Remote Sens. Environ. 114, 725–737. https://doi.org/10.1016/j.rse.2009.11.002

Fidalgo-González, L.A., Arellano-Pérez, S., Álvarez-González, J.G., Castedo-Dorado, F., Ruiz-González, A.D., González-Ferreiro, E., 2019. Estimación de la distribución vertical de combustibles finos del dosel de copas en masas de Pinus sylvestris empleando datos LiDAR de baja densidad. Revista de Teledetección 53, 1. https://doi.org/10.4995/raet.2019.11241

Fischer, R., Knapp, N., Bohn, F., Shugart, H.H., Huth, A., 2019. The Relevance of Forest Structure for Biomass and Productivity in Temperate Forests: New Perspectives for Remote Sensing. Surv. Geophys. 40, 709–734. https://doi.org/10.1007/s10712-019-09519-x

Flynn, W.R.M., Owen, H.J.F., Grieve, S.W.D., Lines, E.R., 2023. Quantifying vegetation indices using terrestrial laser scanning: methodological complexities and ecological insights from a Mediterranean forest. Biogeosciences 20, 2769–2784. https://doi.org/10.5194/bg-20-2769-2023

Forbes, B., Reilly, S., Clark, M., Ferrell, R., Kelly, A., Krause, P., Matley, C., O'Neil, M., Villasenor, M., Disney, M., Wilkes, P., Bentley, L.P., 2022. Comparing Remote Sensing and Field-Based Approaches to Estimate Ladder Fuels and Predict Wildfire Burn Severity. Frontiers in Forests and Global Change 5, 818713. https://doi.org/10.3389/ffgc.2022.818713

García, M., Danson, F.M., Riaño, D., Chuvieco, E., Ramirez, F.A., Bandugula, V., 2011. Terrestrial laser scanning to estimate plot-level forest canopy fuel properties. International Journal of Applied Earth Observation and Geoinformation 13, 636–645. https://doi.org/10.1016/j.jag.2011.03.006

Gough, C.M., Atkins, J.W., Fahey, R.T., Hardiman, B.S., 2019. High rates of primary production in structurally complex forests. Ecology 100, e02864. https://doi.org/10.1002/ecy.2864

Hakkenberg, C.R., Clark, M.L., Bailey, T., Burns, P., Goetz, S.J., 2024. Ladder fuels rather than canopy volumes consistently predict wildfire severity even in extreme topographic-weather conditions. Commun. Earth Environ. 5, 721. https://doi.org/10.1038/s43247-024-01893-8

Hancock, S., Anderson, K., Disney, M., Gaston, K.J., 2017. Measurement of fine-spatial-resolution 3D vegetation structure with airborne waveform lidar: Calibration and validation with voxelised terrestrial lidar. Remote Sens. Environ. 188, 37–50. https://doi.org/10.1016/j.rse.2016.10.041

Hermosilla, T., Ruiz, L.A., Kazakova, A.N., Coops, N.C., Moskal, L.M., 2014. Estimation of forest structure and canopy fuel parameters from small-footprint full-waveform LiDAR data. Int. J. Wildland Fire 23, 224–233. https://doi.org/10.1071/WF13086

Hoffrén, R., Lamelas, M.T., de la Riva, J., 2024. Evaluation of Handheld Mobile Laser Scanner Systems for the Definition of Fuel Types in Structurally Complex Mediterranean Forest Stands. Fire 7, 59. https://doi.org/10.3390/FIRE7020059/S1

Hoffrén, R., Lamelas, M.T., de la Riva, J., Domingo, D., Montealegre, A.L., García-Martín, A., Revilla, S., 2023. Assessing GEDI-NASA system for forest fuels classification using machine learning techniques. International Journal of Applied Earth Observation and Geoinformation 116, 103175. https://doi.org/10.1016/j.jag.2022.103175

Huff, M.H., Ottmar Roger D., Alvarado, E., Vihnanek, R.E., Lehmkuhl, J.F., Hessburg, P.F., Everett, R.L., 1995. Historical and current forest landscapes in eastern Oregon and Washington. Part II: Linking vegetation characteristics to potential fire behavior and related smoke production. (Everett, Richard L., team leader; Eastside forest ecosystem health assessment; Hessburg, Paul F., science team leader and tech. ed., Volume III: assessment.).

Inman, H.F., Bradley, E.L., 1989. The overlapping coefficient as a measure of agreement between probability distributions and point estimation of the overlap of two normal densities. Commun. Stat. Theory Methods 18, 3851–3874. https://doi.org/10.1080/03610928908830127

Kacic, P., Gessner, U., Hakkenberg, C.R., Holzwarth, S., Müller, J., Pierick, K., Seidel, D., Thonfeld, F., Torresani, M., Kuenzer, C., 2025. Characterizing local forest structural complexity based on multi-platform and -sensor derived indicators. Ecol. Indic. 170, 113085. https://doi.org/10.1016/J.ECOLIND.2025.113085

Knuth, K.H., 2019. Optimal data-based binning for histograms and histogram-based probability density models. Digit. Signal Process. 95, 102581. https://doi.org/10.1016/j.dsp.2019.102581

Kramer, H., Collins, B., Kelly, M., Stephens, S., 2014. Quantifying Ladder Fuels: A New Approach Using LiDAR. Forests 5, 1432–1453. https://doi.org/10.3390/f5061432

Kreider, M.R., Higuera, P.E., Parks, S.A., Rice, W.L., White, N., Larson, A.J., 2024. Fire suppression makes wildfires more severe and accentuates impacts of climate change and fuel accumulation. Nat. Commun. 15, 2412. https://doi.org/10.1038/s41467-024-46702-0

Laino, D., Cabo, C., Ordóñez, C., Bolanos, R., Janvier, R., Giulioni, F., Herrmann, M., Hudak, A., Parsons, R., Santin, C., 2026. SegmentedForests: a labelled dataset of terrestrial LiDAR point clouds for semantic segmentation of forests. Forestry: An International Journal of Forest Research 99. https://doi.org/10.1093/forestry/cpaf062

Laino, D., Cabo, C., Prendes, C., Janvier, R., Ordonez, C., Nikonovas, T., Doerr, S., Santin, C., 2024. 3DFin: a software for automated 3D forest inventories from terrestrial point clouds. Forestry: An International Journal of Forest Research 97, 479–496. https://doi.org/10.1093/forestry/cpae020

Little, K., Vitali, R., Belcher, C.M., Kettridge, N., Pellegrini, A.F.A., Ford, A.E.S., Smith, A.M.S., Elliott, A., Voulgarakis, A., Stoof, C.R., Kolden, C.A., Schwilk, D.W., Kennedy, E.B., Newman Thacker, F.E., Millin-Chalabi, G.R., Clay, G.D., Morison, J.I., McCarty, J.L., Ivison, K., Tansey, K., Simpson, K.J., Jones, M.W., Mack, M.C., Fulé, P.Z., Gazzard, R., Harrison, S.P., New, S., Page, S.E., Hall, T.E., Brown, T., Jolly, W.M., Doerr, S., 2025. Priority research directions for wildfire science: Views from a historically fire-prone and an emerging fire-prone country. Philosophical Transactions of the Royal Society B: Biological Sciences 380. https://doi.org/10.1098/RSTB.2024.0001/234873

Liu, J., Zhang, X., Lv, J., Li, X., Du, L., 2025. Gaussian decomposition method for full waveform data of LiDAR base on neural network. Sci. Rep. 15, 5639. https://doi.org/10.1038/s41598-024-82543-z

Liu, X., Feng, Y., Hu, T., Luo, Y., Zhao, X., Wu, J., Maeda, E.E., Ju, W., Liu, L., Guo, Q., Su, Y., 2024. Enhancing ecosystem productivity and stability with increasing canopy structural complexity in global forests. Sci. Adv. 10, 1947. https://doi.org/10.1126/sciadv.adl1947

Macmillan, N.A., Creelman, C.D., 1991. Detection Theory: A User's Guide, 1st edition. ed, Detection Theory: A User's Guide: 2nd edition. Cambridge University Press. https://doi.org/10.4324/9781410611147

Marino, E., Montes, F., Tomé, J.L., Navarro, J.A., Hernando, C., 2018. Vertical forest structure analysis for wildfire prevention: Comparing airborne laser scanning data and stereoscopic hemispherical images. International Journal of Applied Earth Observation and Geoinformation 73, 438–449. https://doi.org/10.1016/j.jag.2018.07.015

Martin-Ducup, O., Dupuy, J.-L., Soma, M., Guerra-Hernandez, J., Marino, E., Fernandes, P.M., Just, A., Corbera, J., Toutchkov, M., Sorribas, C., Bock, J., Piboule, A., Pirotti, F., Pimont, F., 2025. Unlocking the potential of Airborne LiDAR for direct assessment of fuel bulk density and load distributions for wildfire hazard mapping. Agric. For. Meteorol. 362, 110341. https://doi.org/10.1016/j.agrformet.2024.110341

McElhinny, C., Gibbons, P., Brack, C., Bauhus, J., 2005. Forest and woodland stand structural complexity: Its definition and measurement. For. Ecol. Manage. 218, 1–24. https://doi.org/10.1016/j.foreco.2005.08.034

Menning, K.M., Stephens, S.L., 2007. Fire Climbing in the Forest: A Semiqualitative, Semiquantitative Approach to Assessing Ladder Fuel Hazards. Western Journal of Applied Forestry 22, 88–93. https://doi.org/10.1093/wjaf/22.2.88

Moudrý, V., Cord, A.F., Gábor, L., Laurin, G.V., Barták, V., Gdulová, K., Malavasi, M., Rocchini, D., Stereńczak, K., Prošek, J., Klápště, P., Wild, J., 2023. Vegetation structure derived from airborne laser scanning to assess species distribution and habitat suitability: The way forward. Divers. Distrib. 29, 39–50. https://doi.org/10.1111/DDI.13644;JOURNAL:JOURNAL:14724642;WGROUP:STRING:PUBLICATION

Mountrakis, G., Li, Y., 2017. A linearly approximated iterative Gaussian decomposition method for waveform LiDAR processing. ISPRS Journal of Photogrammetry and Remote Sensing 129, 200–211. https://doi.org/10.1016/j.isprsjprs.2017.05.009

Ottmar, R.D., Blake, J.I., Crolly, W.T., 2012. Using fine-scale fuel measurements to assess wildland fuels, potential fire behavior and hazard mitigation treatments in the southeastern USA. For. Ecol. Manage. 273, 1–3. https://doi.org/10.1016/j.foreco.2011.11.003

Ottmar, R.D., Vihnanek, R.E., Wright, C.S., Hudak, A.T., 2014. Ground measurements of fuel and fuel consumption from experimental and operational prescribed fires at Eglin Air Force Base, Florida. In: Wade, Dale D.; Fox, Rebekah L., eds; Robinson, Mikel L., comp. Proceedings of 4th Fire Behavior and Fuels Conference; 18-22 February 2013, Raleigh, NC and 1-4 July 2013, St. Petersburg, Russia. Missoula, MT: International Association of Wildland Fire… 227–234.

Papadopoulos, G.D., Pavlidou, F.-N., 2011. A Comparative Review on Wildfire Simulators. IEEE Syst. J. 5, 233–243. https://doi.org/10.1109/JSYST.2011.2125230

Parks, S.A., Holsinger, L.M., Panunto, M.H., Jolly, W.M., Dobrowski, S.Z., Dillon, G.K., 2018. High-severity fire: evaluating its key drivers and mapping its probability across western US forests. Environmental Research Letters 13, 044037. https://doi.org/10.1088/1748-9326/aab791

Parsons, R., Linn, R., Pimont, F., Hoffman, C., Sauer, J., Winterkamp, J., Sieg, C., Jolly, W., 2017. Numerical Investigation of Aggregated Fuel Spatial Pattern Impacts on Fire Behavior. Land (Basel). 6, 43. https://doi.org/10.3390/land6020043

Pausas, J.G., Fernández-Muñoz, S., 2012. Fire regime changes in the Western Mediterranean Basin: From fuel-limited to drought-driven fire regime. Clim. Change 110, 215–226. https://doi.org/10.1007/S10584-011-0060-6/METRICS

Prichard, S.J., Rowell, E.M., Hudak, A.T., Keane, R.E., Loudermilk, E.L., Lutes, D.C., Ottmar, R.D., Chappell, L.M., Hall, J.A., Hornsby, B.S., 2022. Fuels and consumption. Wildland Fire Smoke in the United States: A Scientific Assessment 11–49. https://doi.org/10.1007/978-3-030-87045-4_2/TABLES/2

Puliti, S., Breidenbach, J., Astrup, R., 2020. Estimation of Forest Growing Stock Volume with UAV Laser Scanning Data: Can It Be Done without Field Data? Remote Sens. (Basel). 12, 1245. https://doi.org/10.3390/rs12081245

Ritter, S.M., Hoffman, C.M., Battaglia, M.A., Linn, R., Mell, W.E., 2023. Vertical and Horizontal Crown Fuel Continuity Influences Group-Scale Ignition and Fuel Consumption. Fire 6, 321. https://doi.org/10.3390/fire6080321

Rocha, K.D., Silva, C.A., Cosenza, D.N., Mohan, M., Klauberg, C., Schlickmann, M.B., Xia, J., Leite, R. V., Almeida, D.R.A. de, Atkins, J.W., Cardil, A., Rowell, E., Parsons, R., Sánchez-López, N., Prichard, S.J., Hudak, A.T., 2023.

Crown-Level Structure and Fuel Load Characterization from Airborne and Terrestrial Laser Scanning in a Longleaf Pine (Pinus palustris Mill.) Forest Ecosystem. Remote Sens. (Basel). 15, 1002. https://doi.org/10.3390/rs15041002

Rodríguez y Silva, F., Guijarro, M., Madrigal, J., Jiménez, E., Molina, J.R., Hernando, C., Vélez, R., Vega, J.A., 2017. Assessment of crown fire initiation and spread models in Mediterranean conifer forests by using data from field and laboratory experiments. For. Syst. 26, e02S. https://doi.org/10.5424/fs/2017262-10652

Rothermel, R.C., 1972. A mathematical model for predicting fire spread in wildland fuels. Res. Pap. INT-115. Ogden, UT: U.S. Department of Agriculture, Intermountain Forest and Range Experiment Station. 40 p. 115.

Rowell, E., Loudermilk, E.L., Hawley, C., Pokswinski, S., Seielstad, C., Queen, Ll., O'Brien, J.J., Hudak, A.T., Goodrick, S., Hiers, J.K., 2020. Coupling terrestrial laser scanning with 3D fuel biomass sampling for advancing wildland fuels characterization. For. Ecol. Manage. 462, 117945. https://doi.org/10.1016/j.foreco.2020.117945

Satapathy, T., Dutta, D., 2025. Characterizing the vertical variability of canopy structure from space: Implications for biodiversity, productivity, and ecosystem functioning. Agric. For. Meteorol. 372, 110710. https://doi.org/10.1016/J.AGRFORMET.2025.110710

Savitzky, A., Golay, M.J.E., 1964. Smoothing and Differentiation of Data by Simplified Least Squares Procedures. Anal. Chem. 36, 1627–1639. https://doi.org/10.1021/AC60214A047/ASSET/AC60214A047.FP.PNG_V03

Scott, J.H., Burgan, R.E., 2005. Standard fire behavior fuel models: A comprehensiveset for use with Rothermel's surface fire spread model.

Simó-Martí, M., Crespo-Peremarch, P., Torralba, J., Ruiz, L.Á., 2025. Analysis of fire behavior in semi-real forest scenarios by integrating tls data into physical simulation models, in: CiGEO Quinto Congreso En Ingeniería Geomática. Editorial Universitat Politècnica de València, Valencia, pp. 1–7. https://doi.org/10.4995/CIGeo2025.2025.19873

Swain, P.H., King, R.C., 1973. Two effective feature selection criteria for multispectral remote sensing, in: The International Joint Conference on Pattern Recognition. LARS Technical Reports, Washington, D.C., pp. 1–6.

Tanase, M.A., Mihai, M.C., Miguel, S., Cantero, A., Tijerin, J., Ruiz-Benito, P., Domingo, D., Garcia-Martin, A., Aponte, C., Lamelas, M.T., 2024. Long-term annual estimation of forest above ground biomass, canopy cover, and height from airborne and spaceborne sensors synergies in the Iberian Peninsula. Environ. Res. 259, 119432. https://doi.org/10.1016/j.envres.2024.119432

Torralba, J., Carbonell-Rivera, J.P., Ruiz, L.Á., Crespo-Peremarch, P., 2022. Analyzing TLS Scan Distribution and Point Density for the Estimation of Forest Stand Structural Parameters. Forests 13, 2115. https://doi.org/10.3390/f13122115

Torralba, J., Crespo-Peremarch, P., Ruiz, L.Á., 2018. Assessing the use of discrete, full-waveform LiDAR and TLS to classify Mediterranean forest species composition. Revista de Teledetección 27. https://doi.org/10.4995/raet.2018.11106

Torralba, J., Crespo-Peremarch, P., Ruiz, L.Á., Carbonell-Rivera, J.P., 2024. Análisis multitemporal de la estructura vertical de ecosistemas forestales mediterráneos mediante datos TLS, in: Teledetección y Cambio Global: Retos y Oportunidades Para Un Crecimiento Azul. XX Congreso de La Asociación Española de Teledetección . pp. 291–294.

Torralba, J., Danson, F.M., Ruiz, L.A., Almonacid-Caballer, J., Crespo-Peremarch, P., Carbonell-Rivera, J.P., 2026. Towards in-field live fuel moisture content estimation using multi-wavelength terrestrial laser scanning. Frontiers in Forests and Global Change 9, 1770225. https://doi.org/10.3389/ffgc.2026.1770225

Viedma, O., Chico, F., Fernández, J.J., Madrigal, C., Safford, H.D., Moreno, J.M., 2020. Disentangling the role of prefire vegetation vs. burning conditions on fire severity in a large forest fire in SE Spain. Remote Sens. Environ. 247, 111891. https://doi.org/10.1016/j.rse.2020.111891

Viedma, O., Moreno, J.M., 2025. Impact of LiDAR pulse density on forest fuels metrics derived using LadderFuelsR. Ecol. Inform. 88, 103135. https://doi.org/10.1016/j.ecoinf.2025.103135

Viedma, O., Silva, C.A., Moreno, J.M., Hudak, A.T., 2024. LadderFuelsR: A new automated tool for vertical fuel continuity analysis and crown base height detection using light detection and ranging. Methods Ecol. Evol. 15, 1958–1967. https://doi.org/10.1111/2041-210X.14427

Wilson, N., Bradstock, R., Bedward, M., 2021. Detecting the effects of logging and wildfire on forest fuel structure using terrestrial laser scanning (TLS). For. Ecol. Manage. 488, 119037. https://doi.org/10.1016/j.foreco.2021.119037

Yépez-Rincón, F.D., Luna-Mendoza, L., Ramírez-Serrato, N.L., Hinojosa-Corona, A., Ferriño-Fierro, A.L., 2021. Assessing vertical structure of an endemic forest in succession using terrestrial laser scanning (TLS). Case study: Guadalupe Island. Remote Sens. Environ. 263, 112563. https://doi.org/10.1016/j.rse.2021.112563

Zhou, T., Popescu, S.C., Krause, K., Sheridan, R.D., Putman, E., 2017. Gold – A novel deconvolution algorithm with optimization for waveform LiDAR processing. ISPRS Journal of Photogrammetry and Remote Sensing 129, 131–150. https://doi.org/10.1016/j.isprsjprs.2017.04.021

Zhuang, W., Mountrakis, G., Wiley, J.J., Beier, C.M., 2015. Estimation of above-ground forest biomass using metrics based on Gaussian decomposition of waveform lidar data. Int. J. Remote Sens. 36, 1871–1889. https://doi.org/10.1080/01431161.2015.1029095

## Appendix A. Additional figures and tables

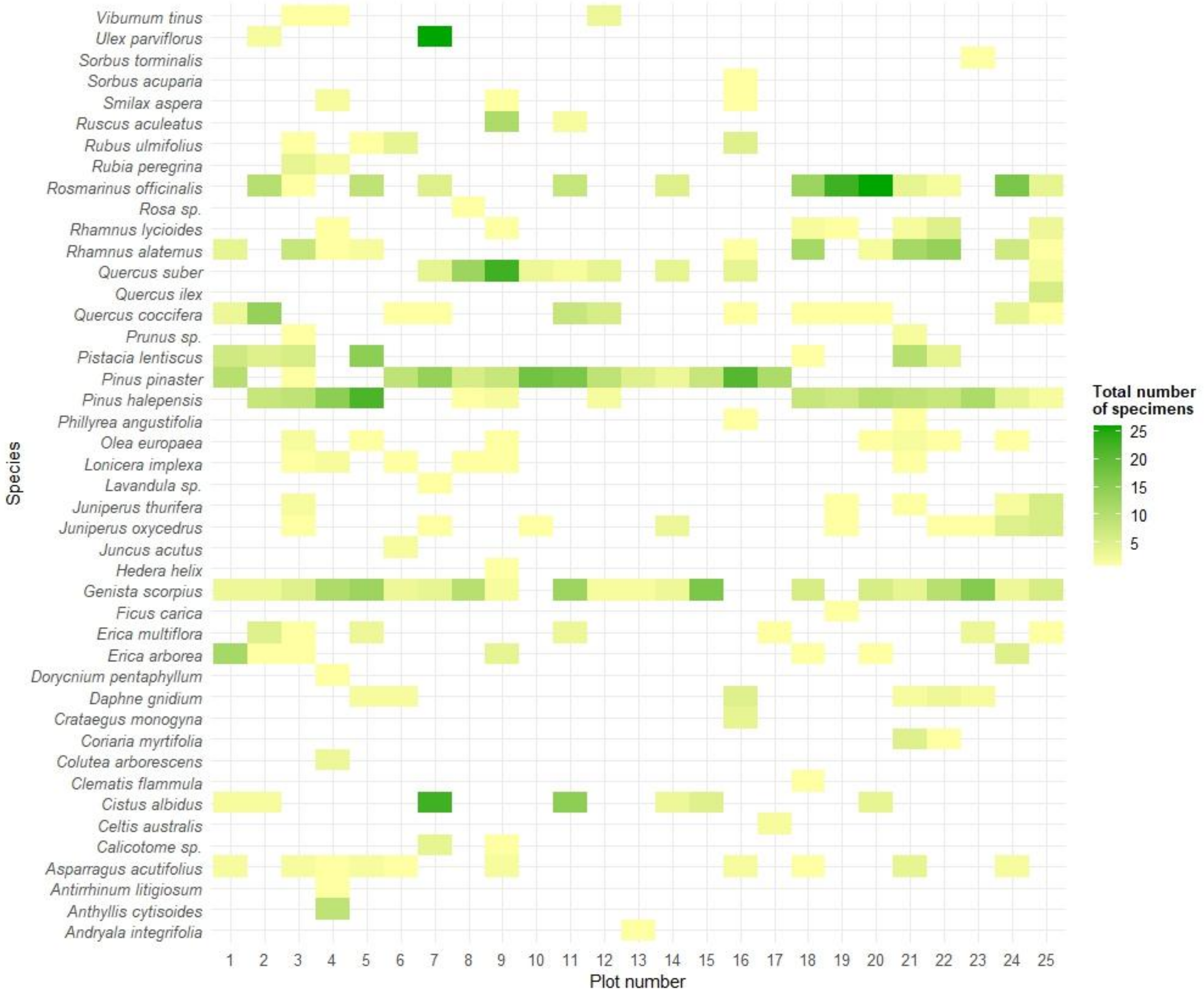

**Figure A1.** Tree and shrub species composition across the 25 study plots. Colors represent the total number of individuals per species within each plot, highlighting differences in species presence and relative abundance among plots.

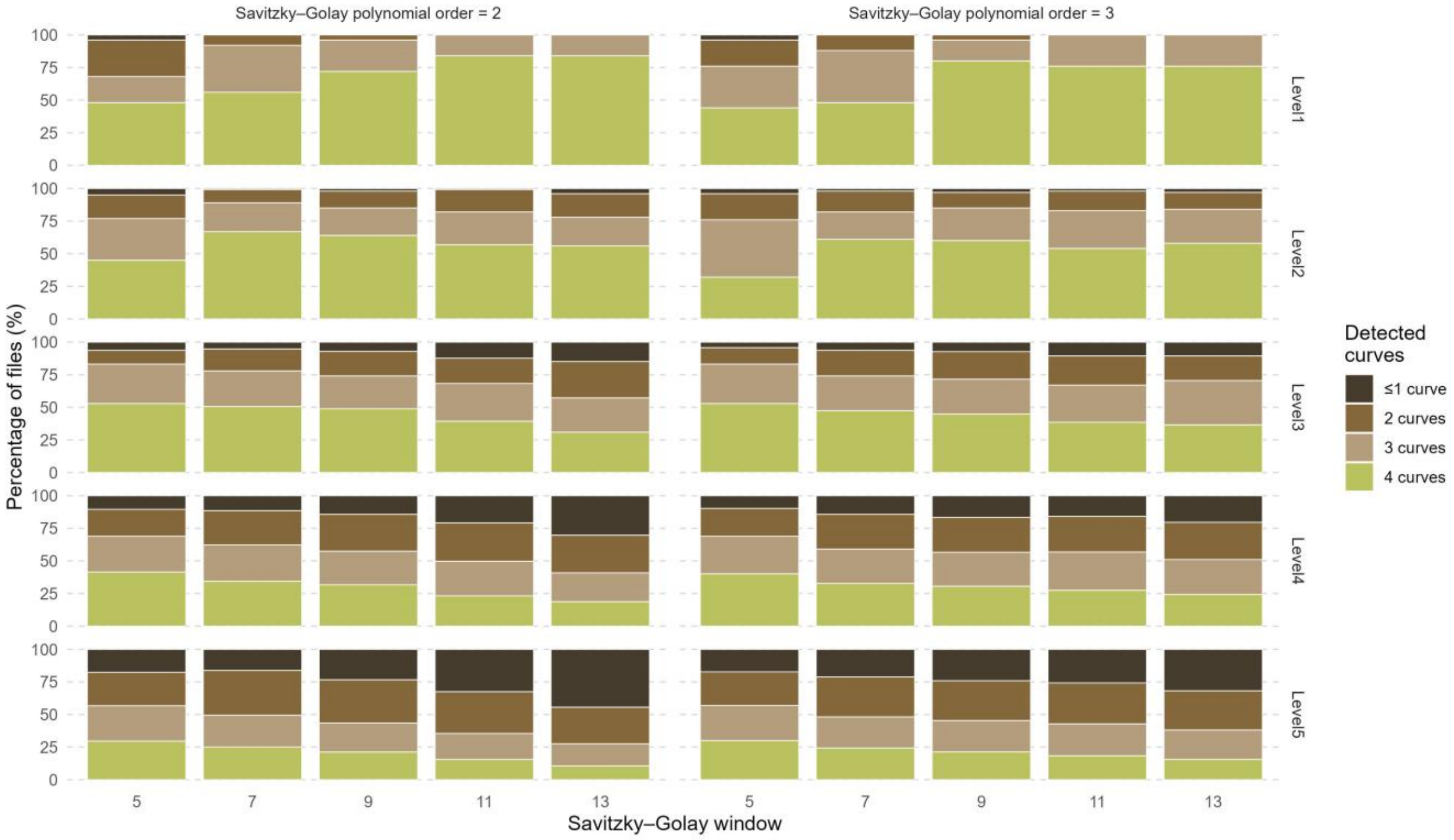

**Figure A2.** Percentage of spatial units in which four, three, two, or one (≤1) Gaussian components were detected as a function of spatial level (Level 1–5), Savitzky-Golay window size ($n$ = 5, 7, 9, 11, and 13), and polynomial order ($p$ = 2 and $p$ = 3). Each bar represents the relative proportion of valid spatial units within each level-window-polynomial combination, expressed as a percentage of the total spatial units that passed the preprocessing and fitting stages. Panels corresponding to each polynomial order allow assessment of the sensitivity of the number of detected components to the degree of smoothing applied. The ≤1 category includes profiles in which only a single Gaussian component was fitted or no curve was obtained.

**Table A1.** Availability of spatial units and Gaussian components across the multiscale processing stages, by analysis level. Theoretical spatial units denote the number generated by the spatial subdivision scheme; available spatial units denote those containing vegetation points after subdivision; and post-binning spatial units denote those retained after vertical discretization. An available shrub component was defined as Gaussian curve 1 (G1) with $\mu \leq 4$ m. The tree component was represented by Gaussian curve 2 (G2) or by a G1 component with $\mu > 4$ m, reclassified as a tree component. In this case, it was omitted from the shrub-tree assignment, and the spatial unit was treated as lacking a shrub component. Available G1-G2 pairs denote spatial units containing both an eligible G1-shrub component and a G2 tree component. Component percentages were calculated relative to the post-binning spatial units.

| Level | Spatial units | | | Gaussian curve | Stratum | Gaussian curve | | |
|---|---|---|---|---|---|---|---|---|
| | Theoretical | Available | Post-binning | | | Available n (%) | Non-available n (%) | Available G1-G2 pairs |
| 1 | 25 | 25 | 25 | 1 | Shrub | 24 (96) | 1 (4) | 24 |
| | | | | 2 | Tree | 25 (100) | 0 (0) | |
| 2 | 100 | 100 | 100 | 1 | Shrub | 82 (82) | 18 (18) | 80 |
| | | | | 2 | Tree | 98 (98) | 2 (2) | |
| 3 | 400 | 399 | 398 | 1 | Shrub | 304 (76) | 94 (24) | 290 |
| | | | | 2 | Tree | 377 (95) | 21 (5) | |
| 4 | 1600 | 1491 | 1475 | 1 | Shrub | 921 (62) | 554 (38) | 847 |
| | | | | 2 | Tree | 1297 (88) | 178 (12) | |
| 5 | 6400 | 5550 | 5355 | 1 | Shrub | 2741 (51) | 2614 (49) | 2588 |
| | | | | 2 | Tree | 4533 (85) | 822 (15) | |

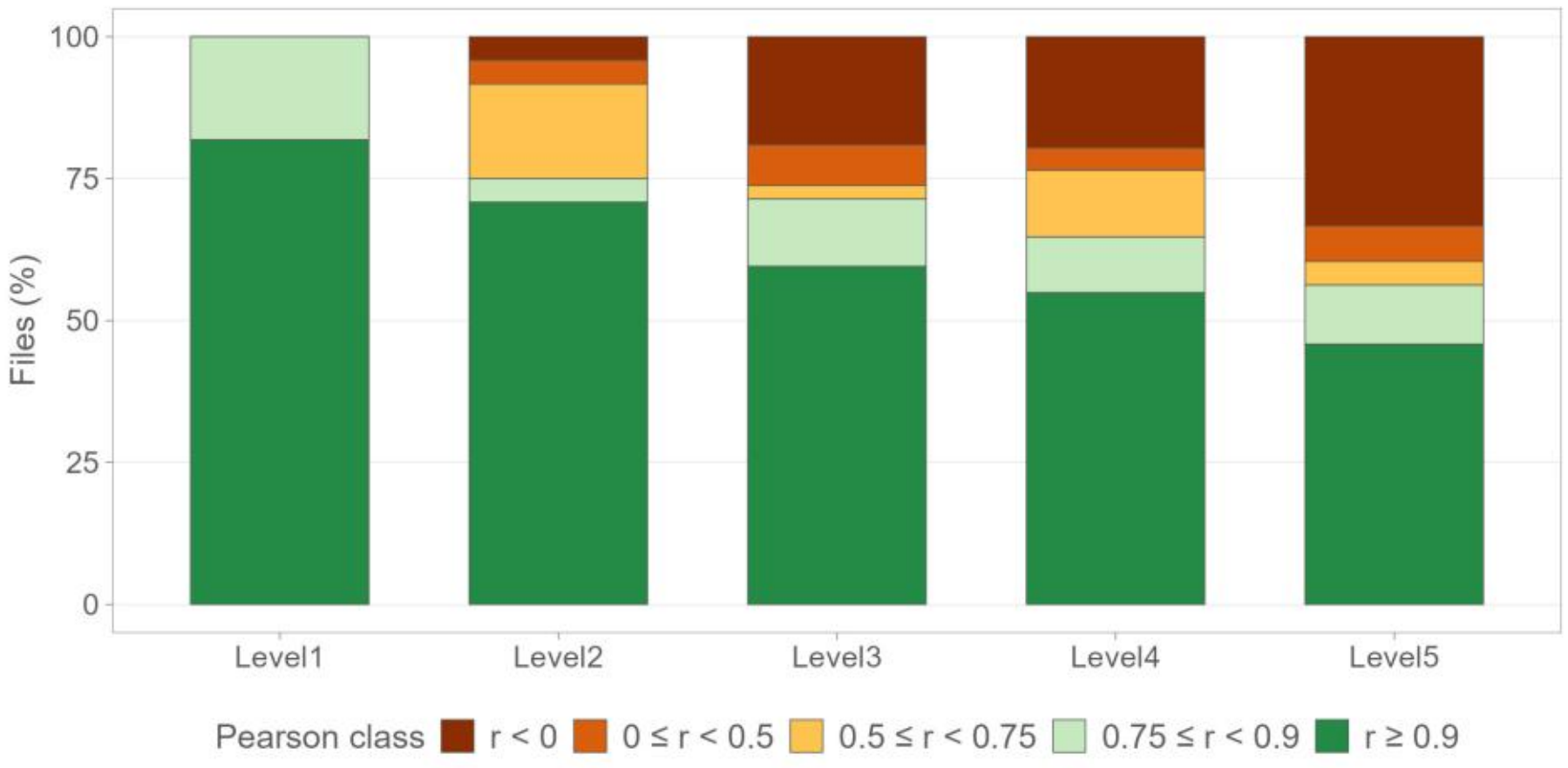

**Figure A3.** Distribution of Pearson correlation values between observed and Gaussian-derived tree-stratum proportions across correlation classes and analysis levels. Paired stacked bars represent the percentage of spatial units falling within each correlation class (negative, weak, moderate, high, and very high correlation). The figure illustrates the increasing variability of correlation values toward finer spatial scales.

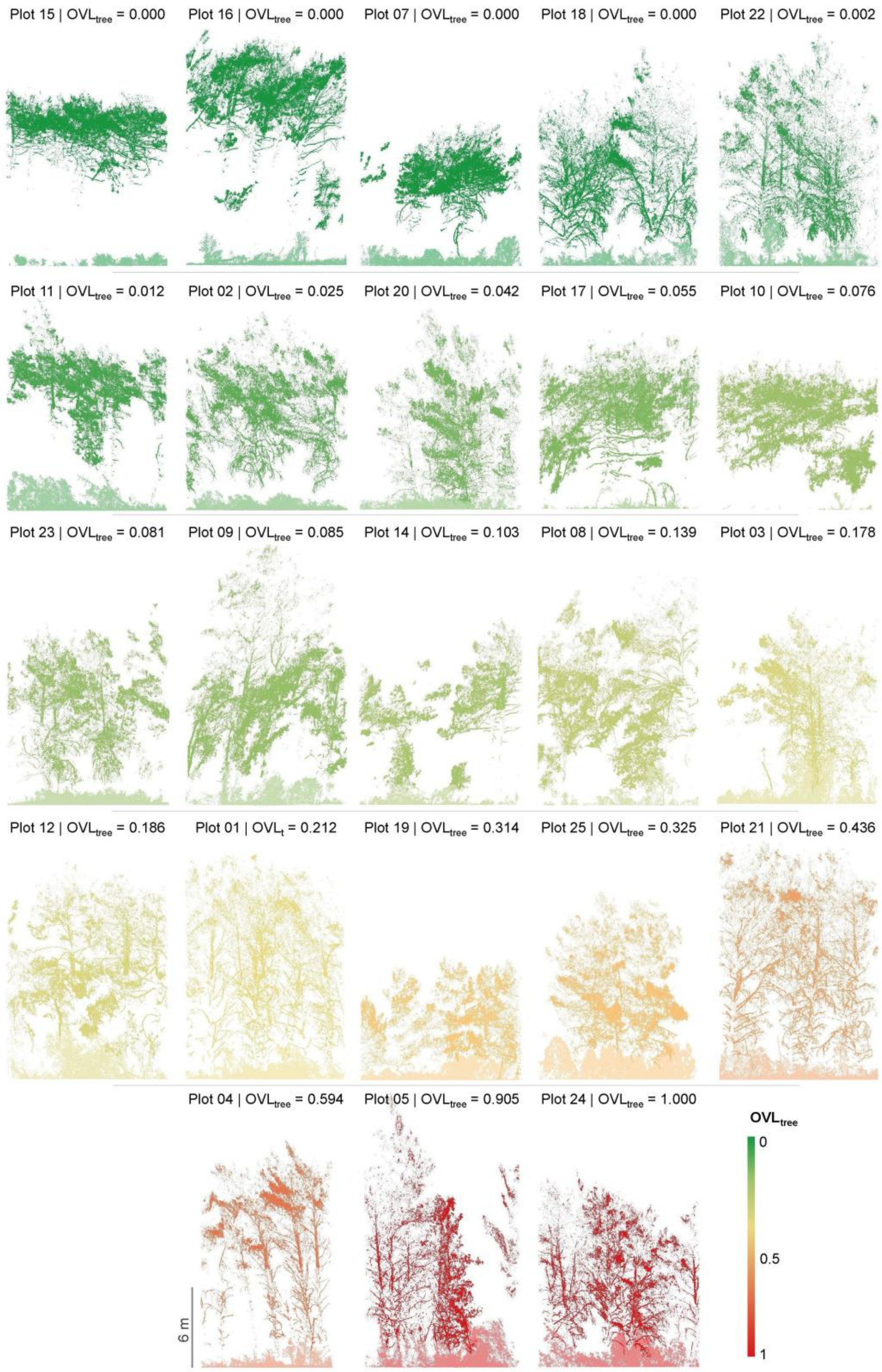

**Figure A4.** TLS point-cloud representation of the 25 study plots at Level 1, colored according to the plot-level directional tree overlap ($OVL_{tree}$). Colors follow a continuous green-yellow-red gradient from low to high $OVL_{tree}$ values. Within each plot, the tree stratum is represented by the color corresponding to its plot-level $OVL_{tree}$ value, and the shrub stratum by a lighter shade of the same color. Plots 06 and 13, which do not have a valid G1-G2 shrub-tree Gaussian pair, are excluded from the figure. $OVL_{tree}$ values are indicated above each plot. The scale bar represents 6 m.

## Appendix B. Initialization of Gaussian-component parameters

For each candidate configuration, the retained local maxima were ordered by increasing bin position. Let $r_i$ denote the bin index of the $i$-th detected peak and $\tilde{y}(r_i)$ the magnitude of the smoothed vertical profile at that position. For a configuration containing $m$ peaks ($m \leq 4$), the initial parameter values supplied to the nonlinear fitting algorithm were defined as:

$$A_i^{(0)} = \frac{2}{3}\tilde{y}(r_i), \qquad i = 1, \dots, m \tag{B1}$$

$$\mu_i^{(0)} = r_i, \qquad i = 1, \dots, m \tag{B2}$$

$$\sigma_1^{(0)} = \frac{r_1}{5} \tag{B3}$$

$$\sigma_i^{(0)} = \frac{r_i - r_{i-1}}{4}, \qquad i = 2, \dots, m \tag{B4}$$

Thus, the initial amplitude of each component was set to two thirds of the corresponding peak magnitude, and its initial mean was placed at the detected peak position. The initial standard deviation of the first component was defined as one fifth of the position of the first peak, whereas the initial standard deviations of the subsequent components were defined as one quarter of the distance between consecutive peaks. These expressions provided only the starting values for the optimization; the final values of $A_i$, $\mu_i$, and $\sigma_i$ were estimated jointly by the Levenberg–Marquardt algorithm.

Initialization and fitting were performed in histogram-bin coordinates. After fitting, the estimated means and standard deviations were converted to height units using the level-specific vertical bin spacing.